\AtBeginDocument{%
  }

\documentclass[sigconf]{acmart}
\usepackage{booktabs}
\usepackage{multirow}
\usepackage{graphicx}
\usepackage{bbding}
\usepackage{threeparttable}
\usepackage{bbm}
\usepackage{bm}
\usepackage{color}
\usepackage{algorithm}
\usepackage{algorithmic}
\usepackage{newfloat}
\usepackage{listings}

\renewcommand\footnotetextcopyrightpermission[1]{}

\begin{document}

\title{Boosting General Trimap-free Matting in the Real-World Image}

\author{Leo Shan}
% \authornote{}
\affiliation{
  \institution{University of Chinese Academy of Sciences}
  \city{Beijing}
  \country{China}}
\email{shanlianlei18@mails.ucas.edu.cn}

\author{Wenzhang Zhou}
\affiliation{%
  \institution{University of Chinese Academy of Sciences}
  %\streetaddress{1 Th{\o}rv{\"a}ld Circle}
  \city{Beijing}
  \country{China}}
\email{zhouwenzhang19@mails.ucas.ac.cn}

\author{Grace Zhao}
\affiliation{
  \institution{University of Chinese Academy of Sciences}
  % \streetaddress{1 Th{\o}rv{\"a}ld Circle}
  \city{Beijing}
  \country{China}}
\email{zhaoguiqin20@mails.ucas.edu.cn}

\renewcommand{\shortauthors}{Leo Shan, Wenzhang Zhou, \& Grace Zhao}

\begin{abstract}
Image matting aims to obtain an alpha matte that separates foreground objects from the background accurately. Recently, trimap-free matting has been well studied because it requires only the original image without any extra input. Such methods usually extract a rough foreground by itself to take place trimap as further guidance. However, the definition of 'foreground' lacks a unified standard and thus ambiguities arise. Besides, the extracted foreground is sometimes incomplete due to inadequate network design. Most importantly, there is not a large-scale real-world matting dataset, and current trimap-free methods trained with synthetic images suffer from large domain shift problems in practice. In this paper, we define the salient object as foreground, which is consistent with human cognition and annotations of the current matting dataset. Meanwhile, data and technologies in salient object detection can be transferred to matting in a breeze. 
To obtain a more accurate and complete alpha matte, we propose a network called \textbf{M}ulti-\textbf{F}eature fusion-based \textbf{C}oarse-to-fine Network \textbf{(MFC-Net)}, which fully integrates multiple features for an accurate and complete alpha matte.
Furthermore, we introduce image harmony in data composition to bridge the gap between synthetic and real images. More importantly, we establish the largest general matting dataset \textbf{(Real-19k)} in the real world to date. Experiments show that our method is significantly effective on both synthetic and real-world images, and the performance in the real-world dataset is far better than existing matting-free methods. Our code and data will be released soon.
\end{abstract}





\maketitle

\section{Introduction}
Digital matting is to accurately extract the foreground object for object-level image composition. It estimates the alpha (opacity) values of the pixels to make an alpha matte for the foreground object.
Most matting methods require a trimap as an auxiliary navigation input \cite{survey}, which contains the clear foreground and background regions, and those unknown regions to be estimated. 
Although trimap is relatively coarse to the alpha matte, the burden of additional input limits the convenience of trimap-based methods in practice, especially for non-interactive applications.
Differently, trimap-free matting methods only need source images. There are two main trimap-free approaches: One replaces the trimap with a background image \cite{backv1,backv2}, and the other generates foreground guidance by itself without any extra input \cite{lfm,attention}.
The former requires static background and movable foreground, which is often unsatisfied in practical applications. Therefore, the latter has more advantages, which only inputs the source image.

\begin{figure}[t]
\centering
\includegraphics[scale=0.25]{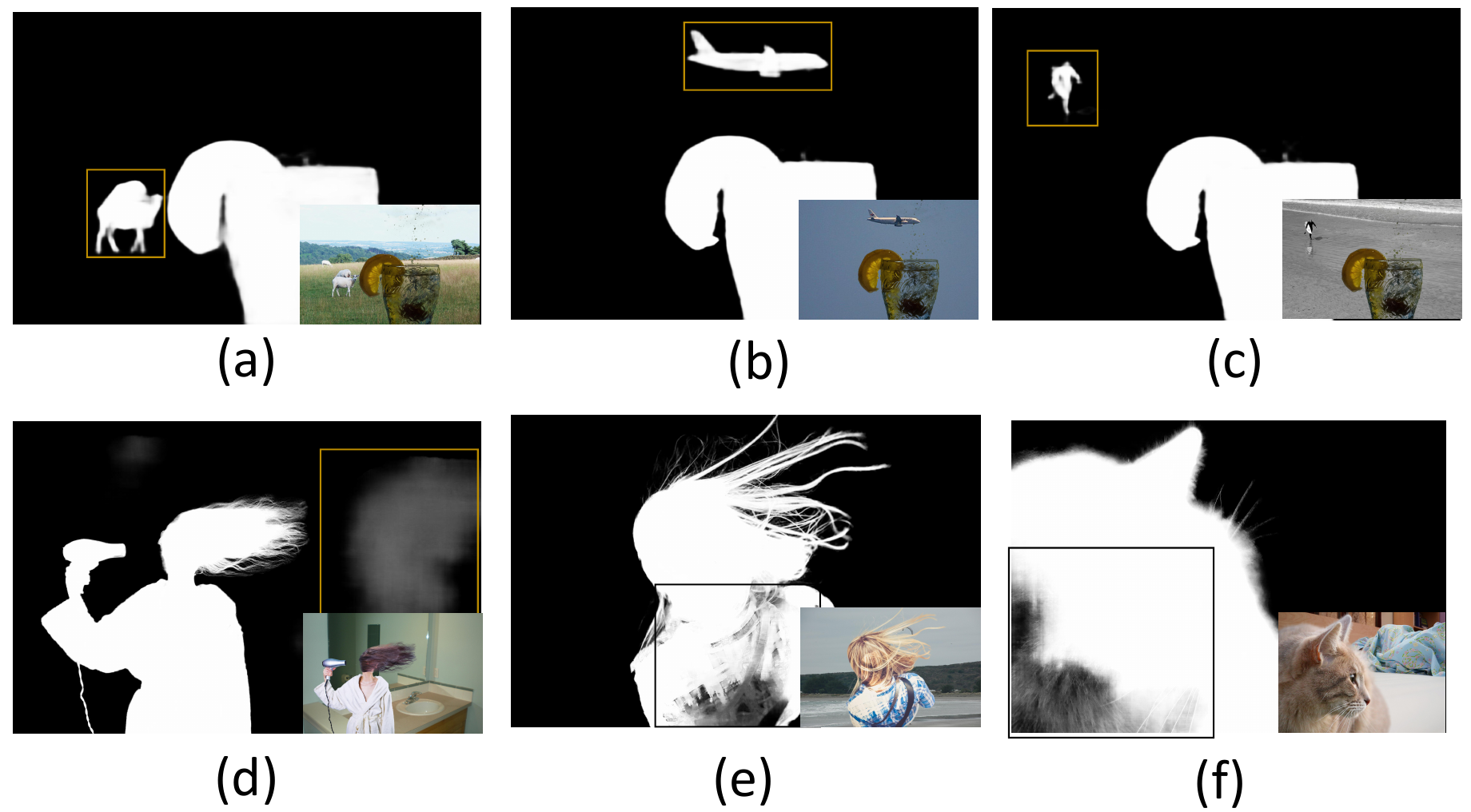}
\caption{
The bad case of existing trimap-free methods for extracting foreground. (a), (b), and (c) extract incorrect objects as the foreground. (d) preserves some visible noise as the foreground. (e) and (f) extract foreground obviously incomplete. (a), (b) and (c) is from modified Deeplab \cite{deeplab} as discussed in \cite{boost}. (d), (e), and (f) are from LFM \cite{lfm}.
}
\label{intro_pic}
\end{figure}

However, without any prior information, the trimap-free methods face many challenges in practical: (1) Definition of the foreground is unclear. As shown in (a), (b), and (c) of Figure \ref{intro_pic}, the plane, goat, and person in the background are misidentified as foreground during training because they are true foregrounds for other images.
(2) The learned foreground is sometimes incomplete, which will cause poor alpha matte \cite{mgmatting}. As shown in (e) (f) of Figure \ref{intro_pic}, some part is missing for a whole foreground.
(3) Using synthetic data for training leads to low generalization for cross-scenes and poor performance in the real world \cite{dim}. Most importantly, there is no large-scale dataset to precisely evaluate the performance of various methods in the real world.
All of these seriously limits the application of trimap-free approaches, and we try to solve the above problems with the following strategies.

\textbf{Giving an explicit foreground definition for matting:}  
Due to the ambiguity of foreground definition, trimap-free methods are usually targeted at a specific category to avoid this confusion, such as human \cite{human2_modnet,human1} or animal \cite{animal}, limiting their versatility.
To solve this problem, we introduce saliency to the foreground and assume the foreground is the salient object in the image, which is a broader and more realistic definition.
This assumption is consistent with human cognition for matting and compatible with the annotation rule of existing matting datasets.
Section I of the supplementary material shows many sample images that support this hypothesis. 
Therefore, techniques and datasets for salient object detection can be naturally migrated to matting, and the annotation of matting can also be semi-automated with the help of saliency detection networks.

\textbf{Improving the completeness of generated foreground:}
The integrity of the foreground has a significant impact on the final matte result, and we guarantee the integrity of the foreground from two aspects: resolution and network structure.
Inspired by GLNet \cite{glnet}, we introduce a ‘coarse to fine’ two-stage training strategy where the foreground is obtained in the coarse module. We use the low-resolution in the coarse module to get holistic high-level information, efficiently reducing false negatives and thus ensuring integrity. Furthermore, we introduce three modules that extract and fully interweave (rather than simply concatenate or add) low-level appearance features, high-level semantic features, and global context features, which further improve completeness. 
The fusion way is distinct from most of the previous works that mainly adopt multiple-level feature integration yet ignore the gap between different features \cite{lfm,attention,u2net}.

\textbf{Promoting the generalization ability in the real world:}  
Most existing matting methods are trained and tested on synthetic datasets, but synthetic images have obvious differences from the real in texture and color \cite{harmony_1}. 
The previous methods achieve good results on synthetic data sets but get a non-neglectable decrease on real-world data sets, especially for the trimap free methods.
To bridge this gap, inspired by image harmony \cite{harmony}, we introduce a data processing method that approximates the synthetic image to the real in texture and brightness. Furthermore, we propose the largest real-world matting dataset to date with 19,083 images. With this, we effectively evaluate the capabilities of various networks in the real world.

To sum up, our contributions are as follows,
\begin{itemize}
\item  We establish the largest real-world matting dataset called Real-19k with 19,083 diverse images to make the precise and objective evaluation. To adapt matting algorithms further to real-world scenes, we novelly introduce image harmony technology in data generation for synthetic datasets.
\item We propose a coarse-to-fine framework for trimap-free matting and model the foreground extraction in the coarse module as saliency detection, which makes the definition of foreground clear and specific. Salient detection methods can be directly introduced to extract and fully interweave multiple features to obtain a complete and accurate foreground.
% \item We firstly introduce harmony technology in synthetic data generation to bridge the performance gap between synthetic and real images and propose a huge matting dataset with 19,083 images to truly test the performance in the real world.
\item Our method achieves SOTA on both synthetic and real-world datasets, and more importantly, our method surpasses all the trimap-free methods in generalization and is robust enough to produce comparable results in multiple scenes.
\end{itemize}

\section{Related Work}

\subsection{Digital Matting}
Matting can be trimap-based or trimap-free. The input of trimap-based network includes a trimap as well as the source image.
In recent years, Convolution Neural Networks (CNNs) have achieved great success in matting.
DIM \cite{dim} created a matting dataset by synthesizing the labeled matte into different background images and trained a deep CNN-based network on this dataset.
IndexNet \cite{indexnet} introduced a new index-guided upsampling and de-pooling operation to better preserve details in prediction.
AlphaGAN \cite{alphagan} introduced a generative adversarial framework to improve the results.
Context-Aware \cite{18_context} proposed a dual encoder and dual decoder structure that simultaneously estimates both foreground and alpha.
GCA \cite{24_gca} further improved performance through context-aware modules.
Background v1 \cite{backv1} and v2 \cite{backv2} used a background image as extra input to get rid of the trimap, but the background image is also difficult to obtain when the foreground object cannot move like mountains or the background changes dynamically. 

A more practical way is to just input the source image, and then the network outputs an exact alpha matte.
Hu et al. in \cite{instance} used instances segmentation network Mask R-CNN \cite{maskrcnn} to produce a coarse output as a trimap, and other trimap-free methods \cite{lfm,attention} also achieve good results, but they can only extract objects of a certain class. 
MODNet \cite{human2_modnet} and Boost\cite{boost} are only for human matting, and AniMatting\cite{animal} is for animal matting, which dramatically limits the range of applications.
LFM \cite{lfm} and HAttMatting \cite{attention} are used for general matting and works well on synthetic images but perform poorly in real-world images, mainly because of the incomplete prediction of foreground, especially in some scenarios not included in the training set.

MODNet \cite{modnet} optimizes a series of sub-objectives simultaneously with explicit constraints. In addition, it introduces a spatial pyramid pooling module to fuse multi-scale features for semantic estimation.
MatteFormer \cite{matteformer} proposes a transformer-based image matting model.
The starting point of AIM \cite{aim} is similar to ours, which is also to extend the matting task to natural images with salient transparent/detailed foreground or non-salient foreground, but it gets poor performance due to imperfect network structure and training strategy.

Our work retains the advantage of not requiring trimap and is not limited to only a single category, and more importantly, our work has achieved competitive performance on both synthetic and real-world images.

\subsection{Salient Object Detection}
Salient Object Detection (SOD) can be seen as an subtask of segmentation \cite{shan2021class,shan2021decouple,shan2021densenet,shan2021uhrsnet,shan2022class,shan2022mbnet,shan2023boosting,shan2023data,shan2023incremental,wu2023continual}.
SOD aims to explore regions more attentive than the surrounding areas on images, at which point it is similar to foreground extraction in trimap-free matting. There are two main challenges for SOD, namely scale variance and unknown object category in inference. Recent works use the different fusions of multi-scale features to solve these problems. F3Net \cite{f3net} proposed a cross-feature module and cascaded feedback decoder with a pixel position-aware minimizing loss. MINet \cite{minet} integrated multi-scale features to obtain more efficient representation and embedded self-interaction modules in each decoder unit. GCPANet \cite{gcpanet} further learned more comprehensive features, emphasizing the information difference between different layers. The development in SOD provides inspiration for foreground extraction, and we follow the mind of SOD but do some simplification to adapt the matting task.
In our main work, the coarse module comes from GCPANet. In fact, any salient object detection network can be used as the coarse module.

\section{The Released Real-World Dataset}
\begin{figure*}[htbp]
\centering
\includegraphics[scale=0.236]{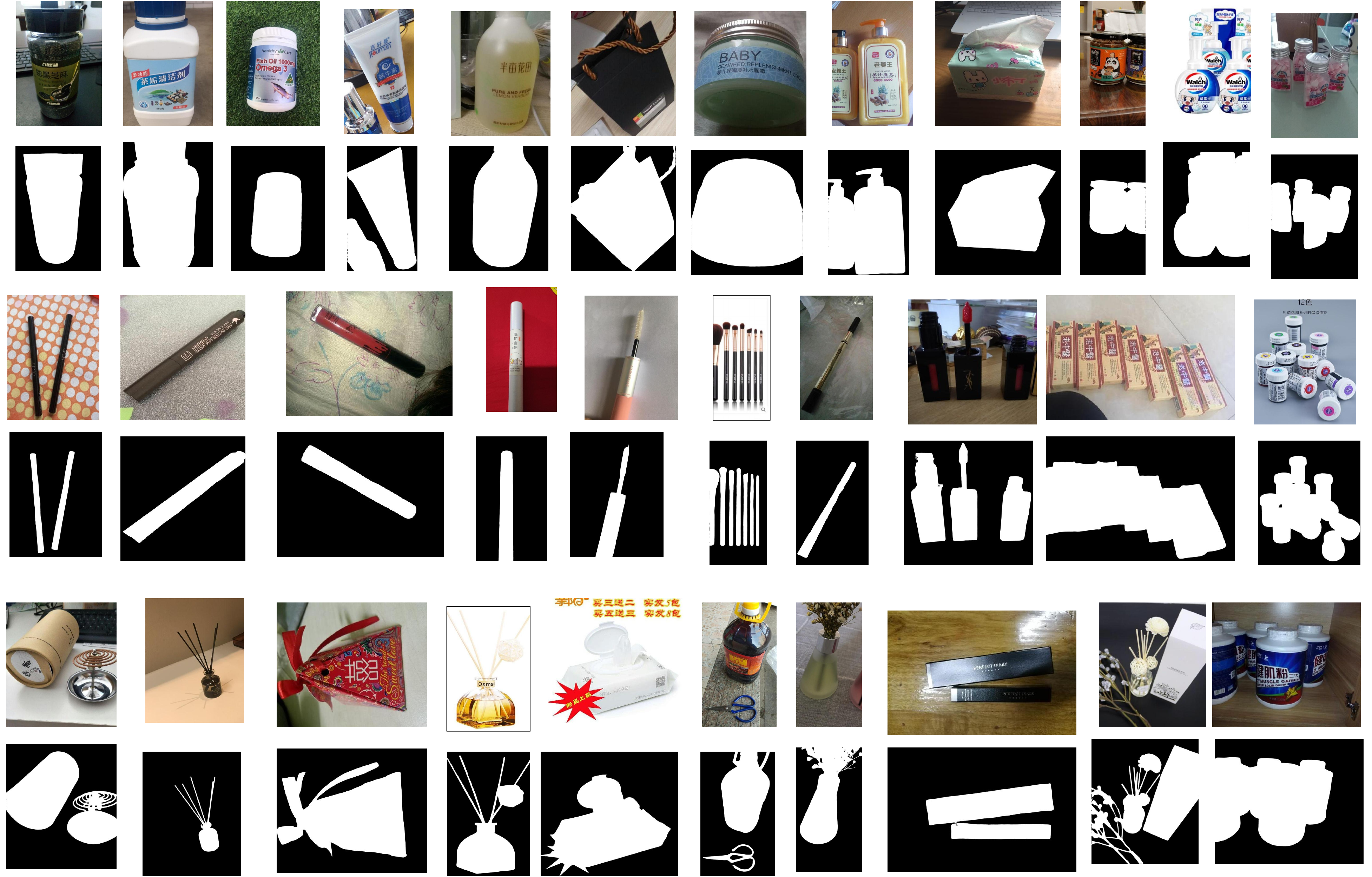}
\caption{
Sample of the proposed Real-19k.
Note that the annotation is alpha-level.
}
\label{dataset_pic}
\end{figure*}

\begin{table}[!htbp]
\centering
  \caption{Class and image numbers of real-world matting datasets.}
  \label{data_number}
    \scalebox{1.1}{
   \begin{tabular}{c|c|c}
    \toprule
     Dataset&Classes&Amount\\
\midrule
Composition-1k \cite{dim}&-&481\\
Distinction-646 \cite{attention}&20&646\\
Semantic \cite{semantic}&20&815\\
AIM-dataset \cite{aim}&-&500\\
PPM-100 \cite{modnet}&-&100\\
\midrule
\textbf{Real-19k} &100+&\textbf{19,083}\\
  \bottomrule
\end{tabular}
}
\end{table}

\textbf{Publicly available datasets:}
The existing synthetic datasets have extensively promoted the development of matting. Nevertheless, the image number of these datasets is limited, object categories are not rich, and more critically, synthesized images look different from real-world images.
Except for synthetic datasets Composition-1k \cite{dim} and Distinct-646 \cite{attention}, people have released some real-world data sets recently, but only for portrait matting.
Liu et al. in \cite{boost} released a large-scale portrait matting dataset with mixed coarse and fine annotations.
MGMatting \cite{mgmatting} published a real-world portrait dataset for the test.
Lin and Sun \cite{backv2,video_matting_dataset} proposed high-resolution video portrait matting datasets.
These matting datasets focused on portrait matting in videos and for remote video conferencing or virtual face swapping. 
However, matting technology is not limited to video communication that requires portrait matting but also has huge application scenarios in the design of product posters, etc. In the scenes, there are many categories of objects.

\textbf{Real-19k:}
To better evaluate the matting methods in the real-world scenario,  we establish a real-world matting dataset named \textbf{Real-19k}. We collect 19,083 diverse images and label them by experts, giving annotation of the detailed alpha matte and salient foreground. Real-19k covers various items, including toiletry, snack, toy, food, medicine, transportation, etc. Its sample number also greatly exceeds the existing datasets, as shown in Table \ref{data_number}. Besides, we split Real-19K into three parts for ease of training and testing. The number of training, verification, and testing sets are 2000, 2000, and 15083, respectively. A larger test set can adequately validate the performance of the model in the real world.
% More detailed descriptions and sample illustrations are given in Section II of the supplementary material.
Samples of the proposed Real-19k are shown in Figure \ref{dataset_pic}, with categories covering all commodities in daily use.
All foreground objects are the salient object in the image, and all annotations are alpha-level compliant with matting standards.
% Our data set mainly targets at the matting of general objects and covers various categories, and the application scenario is the generation of promotional posters of commodities, etc.

% \vspace{-0.2cm}

\section{Methodology}
\label{structure_sec}
\begin{figure*}[htbp]
\centering
\includegraphics[scale=0.60]{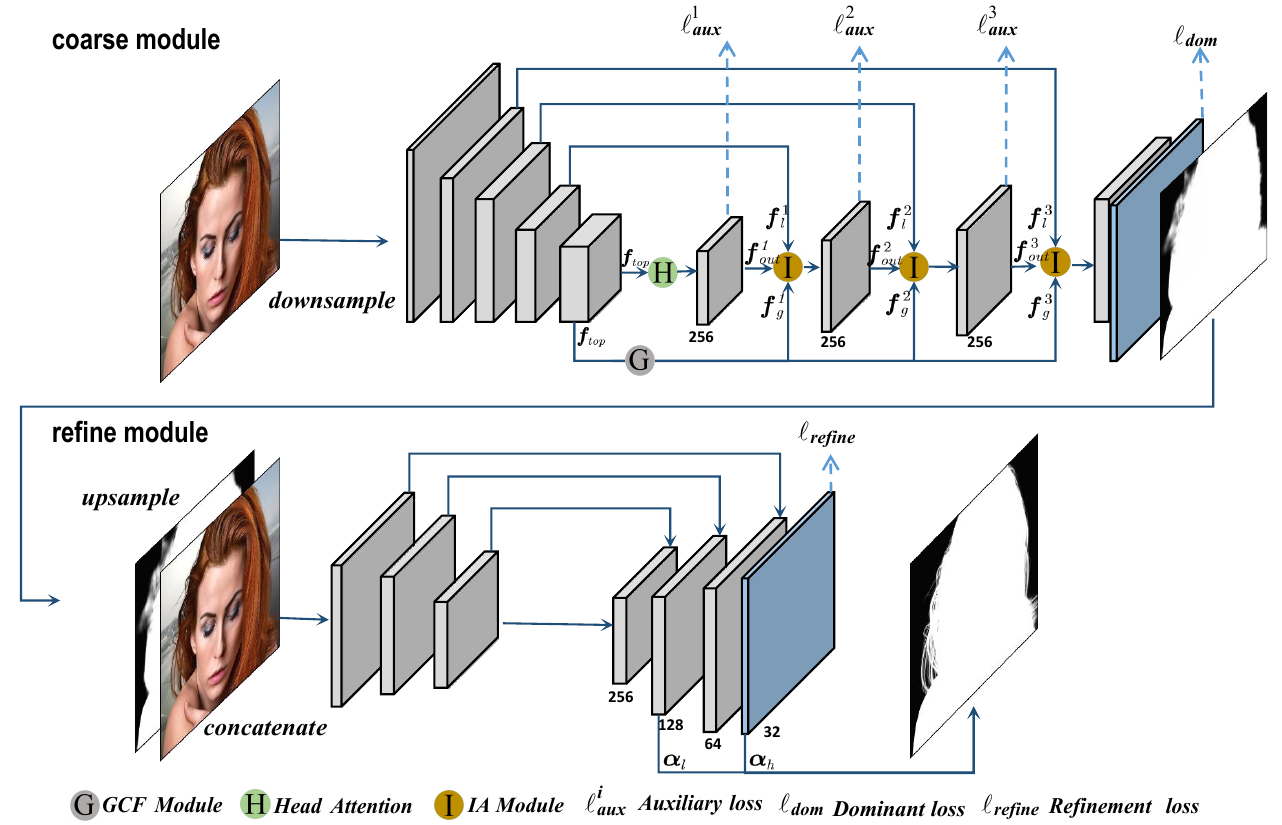}
\caption{
Overview of the MFC-Net.
The top shows the coarse module, which inputs the downsampled image and outputs the corresponding foreground. The bottom illustrates the refine module, which inputs upsampled foreground combined with the original image and outputs the alpha-level matting result. Numbers under the feature maps(like 256) are the numbers of channels.
}
\label{pic_overview}
\end{figure*}
% \subsection{Overview}

%In this section, we first introduce the network consisting of the coarse and refine module and then the data generation method with image harmony.

\subsection{Overview}
In this section, we first introduce our Multi-Feature fusion based Coarse-to-fine Network (MFC-Net), then the data generation method with image harmony.
Our framework adopts a two-stage coarse-to-fine approach. First, we use a  coarse module to extract the foreground from the downsampled source image. Then, the obtained foreground and the original image are sent to a refine module to generate the final alpha matte. Both modules adopt the encoder-decoder structure \cite{unet}, and we make some adjustments for the matting task. The entire framework is shown in Figure \ref{pic_overview}.
Finally, we introduce image harmony to make the synthetic data look real, thus enhancing the generalization of the model in the real world.

\subsection{Coarse Module}
The coarse module can be any salient object detection network.
% Here we mainly learn from GCPANet \cite{gcpanet} and make target improvements.
We inherit the overall architecture of GCPANet \cite{gcpanet}, but do some targeted simplification and improvement to make it more suitable for matting.
To fully utilize different-level features and thus extract more complete and accurate foregrounds, we inherit and improve \textbf{H}ead \textbf{A}ttention (HA), \textbf{G}lobal \textbf{C}ontext \textbf{F}low (GCF) and \textbf{I}nter-weaved \textbf{A}ggregation (IA) in coarse module.

HA is to enhance the spatial regions and feature channels with a high response to foreground objects to generate the first-stage high-level features. GCF produces the global context information, which captures the general information among different foreground regions.
IA fuses the low-level appearance information (from skip connection), high-level semantic information (from HA), and global context information (from GCF) three times in an interweaved way, by which progressively learn more discriminative features for a complete foreground (salient mask). We describe the specific architecture in the following.

\textbf{HA and GCF:}
Since the top layer features are redundant (the feature dimension of ResNet-50 \cite{resnet} is 2048) for foreground extraction, 
we introduce HA following the top layer to learn more representative and discriminative features as well as reducing the channel dimension.
$\boldsymbol{f}_{top}$ is the top (final) layer of ResNet and is the input of HA.
Similar to conditional convolution \cite{solov1,gcpanet}, we put $\boldsymbol{f}_{top}$ in a convolution layer and then split the output in channel dimension to get $\boldsymbol{W}$ and $\boldsymbol{b}$, which can be regarded as the attention maps. The original $\boldsymbol{f}_{top}$ is then multiplied by $\boldsymbol{W}$ and added to $\boldsymbol{b}$ to get the final output. 
Compared with attentions in GCPANet \cite{gcpanet} and segmentation \cite{condinst,solov1}, the proposed HA only conduct three operations.
The specific process is shown in the following,
% \vspace{-0.2cm}
\begin{align}
&\boldsymbol{W},\boldsymbol{b}=split \circ conv_{1} 
\left(\boldsymbol{f}_{top}\right),
\end{align}
\begin{align}
&\widehat{\boldsymbol{f}}=\delta \circ conv_{2} \left(\boldsymbol{f}_{top}\right),
\end{align}
\begin{align}
&\boldsymbol{f}_{out}^{1}=\delta(\boldsymbol{W} \odot \widehat{\boldsymbol{f}}+\boldsymbol{b}),
\end{align}
where $\circ$ denotes function composition, $\delta$ denotes the ReLU activation function, and $\odot$ denotes element-wise multiplication.
The output channels of $conv_{1}$ and $conv_{2}$ are $512$ and $256$ respectively. $split$ operator means to split on the channel dimension equally, so that the output $\boldsymbol{W}$ and $\boldsymbol{b}$ can both multiply and then add directly.

\begin{figure}[htbp]
\centering
\includegraphics[scale=0.6]{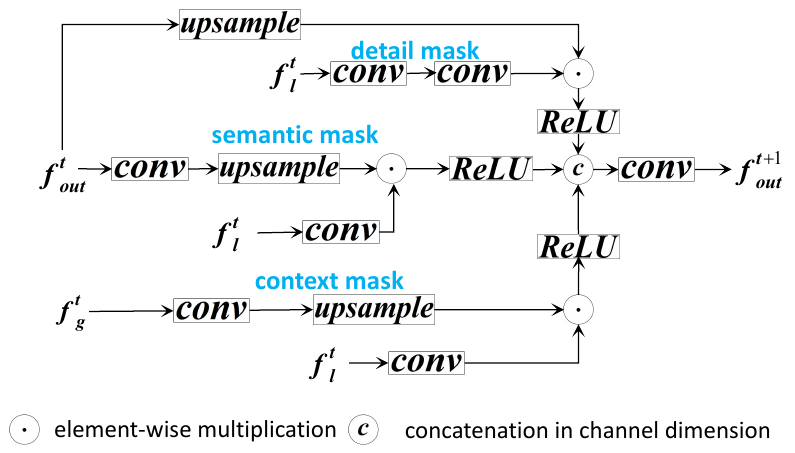}
\label{pic_fia}
\caption{
Overview of the IA.
The input is $\boldsymbol{f}_{out}^{t}$, $\boldsymbol{f}_{g}^{t}$, and $\boldsymbol{f}_{l}^{t}$.
$\boldsymbol{f}_{l}^{t}$ is the low-level appearance feature from encoder module via skip connection.
$\boldsymbol{f}_{out}^{t}$ is high-level semantic feature and is the output of the HA or the previous IA.
$\boldsymbol{f}_{g}^{t}$ is global context information which is the output of GCF. $t$ represents which stage it is, there are three stages as shown in Figure \ref{pic_overview}.
}
\label{pic_fia}
\end{figure}

GCF module is to capture the global context information embedded into the IA module at each stage. 
Firstly, global average pooling \cite{gap_chen} is used to obtain global context information, and then different weights are reassigned to different channels.
% -----------------
% More specifically, for each stage, the process can be described as,
% \begin{align}
% &\boldsymbol{f}_{gap}=averpool\left( \boldsymbol{f}_{top} \right) \\
% &\boldsymbol{y}^{t}=\sigma \circ fc^{t}\left(\boldsymbol{f}_{gap}\right) \\
% &\tilde{\boldsymbol{f}}^{t}=\delta \circ {conv}^{t}\left(\boldsymbol{f}_{top}\right)\\
% &\boldsymbol{f}_{g}^{t}=\tilde{\boldsymbol{f}}^{t} \odot \boldsymbol{y}^{t}
% \end{align}
% where $\boldsymbol{f}_{gap}$ is obatained by $\boldsymbol{f}_{top}$ through a global average pooling layer ($averpool$). 
% $ \boldsymbol{f}_{top} \in {R}^{h \times w \times c}$, $ \boldsymbol{f}_{gap} \in {R}^{1 \times 1 \times c}$, $\boldsymbol{y}^{t} \in {R}^{1 \times 1 \times 256}$, and $\tilde{\boldsymbol{f}}^{t} \in {R}^{h \times w \times 256}$.
% where $fc(.)$ denotes Full-Connected (FC) layers, and $\sigma$ is the sigmoid operation.
% $\boldsymbol{f}_{g}^{t}$ is the output of GCF module, which includes complete global context information.
% $t$ represents which stage it is, there are three stages as shown in Figure \ref{pic_overview}.
% -----------------

\textbf{IA:}
Generally, low-level features contain more details but also more background noises; meanwhile, high-level features contain more abstract semantic information, which is beneficial to locate the foreground and suppress the noises. In addition, the global context information is useful to generate more complete and accurate foreground. Thus we propose IA module to combine these three features effectively and give full play to their respective advantages.
Instead of commonly used concatenation or addition operation, we use a more efficient operation, multiplication.
The specific operation of the proposed IA is shown in Figure \ref{pic_fia}.
From top to bottom in the diagram, we get detail, semantic and context information in turn. Compared with GCPANet \cite{gcpanet}, the structure of each part of our IA is clearer and each role is more specific.

% \vspace{-0.2cm}
\textbf{Loss of coarse module:}
We use the common Binary Cross-Entropy (BCE) as the loss, as shown, 
\begin{equation}
\ell =-\frac{1}{N} \sum_{i=1}^{H} \sum_{j=1}^{W}\left[p_{ij} \log \left(\hat{p}_{ij}\right)\right.
\left.+\left(1-p_{i j}\right) \log \left(1-\hat{p}_{ij}\right)\right]\\
\end{equation}
where $\hat{p}_{ij}$ is the prediction and ${p}_{ij}$ is the ground truth.
$H$ and $W$ are the height and width of the image, and $N=H \times W$.
We add BCE loss at several positions in coarse module, one of them is after the predicted coarse mask which is denoted as $\ell_{\text{dom}}$, while the rest are shown in Figure \ref{pic_overview} and denoted as $\ell_{\text {aux}}$. We combines all these losses shown as follows, 
\begin{equation}
\ell_{\text {coarse}}=\ell_{\text {dom }}+\sum_{i=1}^{3} \lambda_{i} \ell_{\text {aux}}^{i}
\end{equation}
where $\lambda$ represents the weight of $\ell_{\text {aux}}$, which are $0.8$, $0.6$, and $0.4$, respectively.
The intermediate features go through a $3 \times 3$ convolution to make the channel dimension to $1$, and then the output and the label are used to compute $\ell_{\text {aux}}$.
The branches to auxiliary loss only exist during the training stage, and they will be abandoned when inference.

\subsection{Refine Module}
An overview of the refine module is shown at the bottom of Figure \ref{pic_overview}. The architecture follows the popular encoder-decoder network with skip connections. Refine module takes an image and a coarse foreground mask as input. During decoding, similar to previous networks in SOD \cite{egnet}, the module has a extra side output at each feature level.
Specifically, the intermediate feature goes through two $3 \times 3$ convolutions to reduce the channel dimension to $32$ and then to $1$.

Through the visual observation of the results, we found the suppression effect of the background noise is better in the low-resolution output, while the edge part of the high-resolution output is more accurate.
Linear fusion of intermediate outputs or only selecting the final output cannot make full use of the outputs with different resolutions.
Based on this, we use high-resolution output $\boldsymbol{\alpha}_{h}$ for the edge region and low-resolution $\boldsymbol{\alpha}_{l}$ for the rest. The details are shown as follows,
\begin{equation}
\boldsymbol{\hat{\alpha}}=\boldsymbol{g} \odot \boldsymbol{\alpha}_{h} +(1-\boldsymbol{g}) \odot  upsample \odot \boldsymbol{\alpha}_{l}
\end{equation}
\begin{equation}
\boldsymbol{g}=f_{quant}(\boldsymbol{\alpha_{h}})
\label{eq_11}
\end{equation}
\begin{equation}
f_{quant}=\left\{\begin{array}{ll}
1 &  { if  \quad 0<\alpha_{h}(x, y)<1} \\
0 &   otherwise
\end{array}\right.
\end{equation}
The location of $\boldsymbol{\alpha}_{l}$ and $\boldsymbol{\alpha}_{h}$ is shown in Figure \ref{pic_overview}, and $\boldsymbol{g}$ represents the edge (unknown) part.

\textbf{Loss of refine module:}
We use $l_{1}$ regression loss, composition loss \cite{18_context} and Laplacian loss \cite{dim} for the refine module, and they are respectively expressed as $\mathcal{\ell}_{l1}$, $\mathcal{\ell}_{comp}$ and $\mathcal{\ell}_{lap}$.
To make the training focus more on the edge area, we modulate the loss with $\boldsymbol{g}$. After this operation, only the unknown area is optimized during the training process so that the optimization of the two modules are different and complementary, thus the $\mathcal{\ell}_{refine}$ is defined as,
\begin{equation}
\mathcal{\ell}_{refine}
=\mathcal{\ell}_{l1}(\boldsymbol{\hat{{\alpha}_{u}}}, \boldsymbol{\alpha}_{u})
+\mathcal{\ell}_{comp}(\boldsymbol{\hat{{\alpha}_{u}}}, \boldsymbol{\alpha}_{u})
+\mathcal{\ell}_{lap}(\boldsymbol{\hat{\alpha}_{u}}, \boldsymbol{\alpha}_{u})
\end{equation}
where $\boldsymbol{\hat{{\alpha}_{u}}}=\boldsymbol{\hat{\alpha}} \odot \boldsymbol{g}$, $\boldsymbol{{\alpha}_{u}}=\boldsymbol{\alpha} \odot \boldsymbol{g}$,
$\boldsymbol{\hat{\alpha}}$ is the output of refine module, and $\boldsymbol{\alpha}$ is the ground truth.
$\boldsymbol{g}$ is calculated by Eq. (\ref{eq_11}).

\subsection{Image Harmony}
\label{generation}
Image harmony is about making the synthetic image look more realistic \cite{harmony_1}. Image harmony can be regarded as a foreground-to-background style conversion problem \cite{harmony}, and we render the foreground image to maintain a visual style similar to the background image. Stylistic guidance in the background is very important because the foreground image needs to be transformed into a different look when pasted into a different background image. In order to produce a consistent and realistic composite image, we expect a uniform transfer operation that adaptively adjusts the style of the foreground object to maintain perfect harmony with the new background image captured in different environments. To this end,  we put the synthetic image passes through a U-Net \cite{unet}, and a learnable norm layer is proposed in the encoder part, which aligns the mean and variance of the channel level activated by the foreground to match the mean and variance learned from the background.
To the best of our knows, we are the first to use image harmony (style consistency) in data synthesis for the training of matting networks and make great success in bridging the performance gap between the synthetic and real-world images. 

Specifically, the input of the normalization module consists of two parts, i.e., the mask of foreground $M$ and the features of the encoder layer $F$. Take the module in the $i$-th layer for example. Let $F^{i} \in \mathbb{R}^{H^{i} \times W^{i} \times C^{i}}$ and $M^{i} \in \mathbb{R}^{H^{i} \times W^{i}}$ be the resized foreground mask in the $i$-th layer, where $H^{i}, W^{i}, C^{i}$ denote the height, width, and number of channels of feature $F^{i}$, respectively. 
\begin{figure}[htbp]
\centering
\includegraphics[scale=0.33]{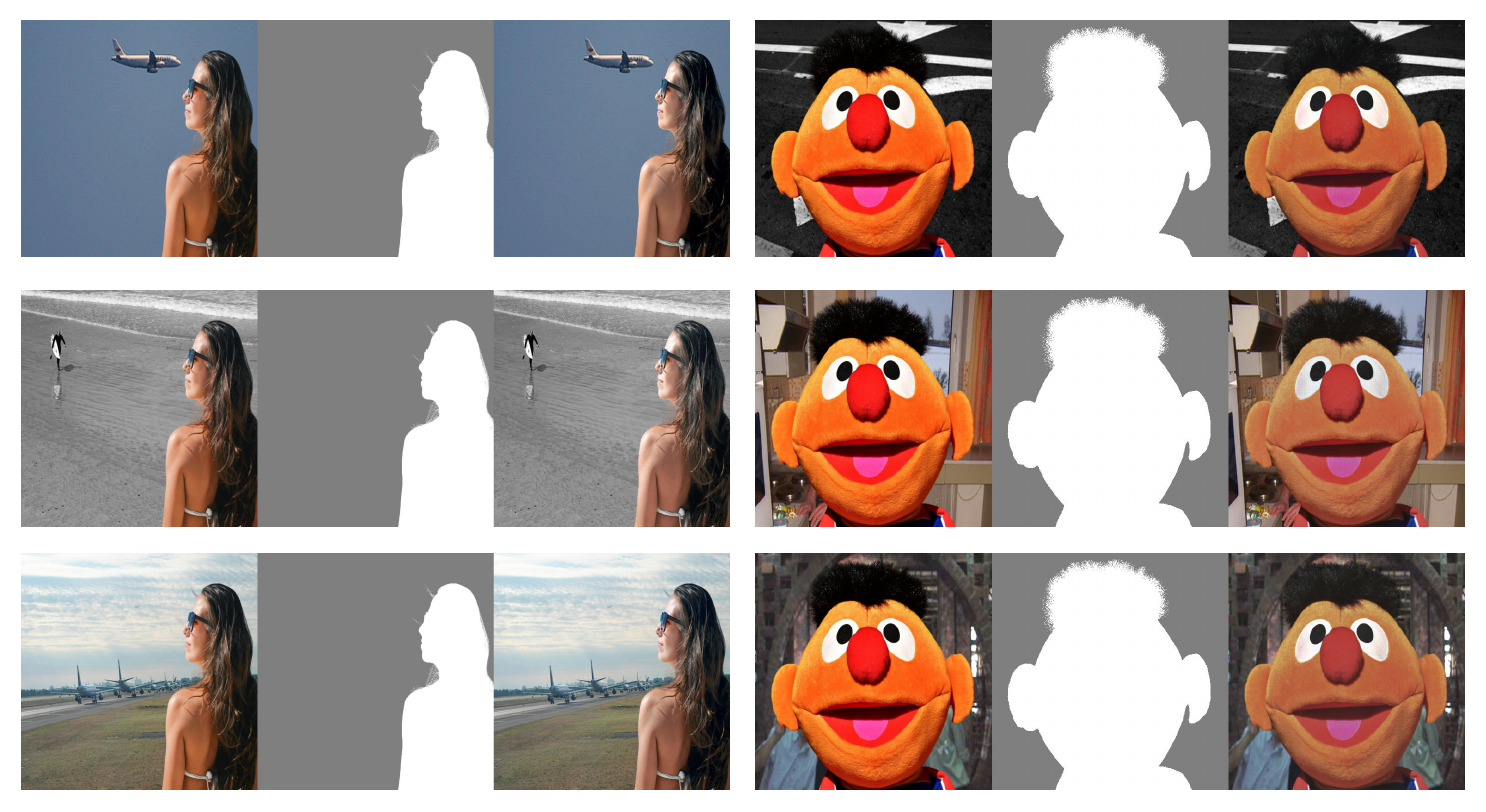}
\caption{
The comparison between synthetic images without/with harmony. 
The left of the mask is without harmony, and the right is with harmony.
It can be seen that 'harmony' adjusts the texture and brightness of the foreground, which adapts to different backgrounds, thereby making the synthetic image more natural.
}
\label{harmony_pic}
\end{figure}
We first multiply the input features $F^{i}$ by the foreground mask and its corresponding background mask. Then we normalize the foreground features by IN \cite{ins_norm}, and then affine the normalized features with leaned scale and bias from the background features. The new activation value $\bar{F}^{i}$ at $(h, w, c)$ in the foreground region is computed by:
\begin{equation}
\bar{F}_{h, w, c}=\gamma_{c}^{i} \frac{F_{h, w, c}^{i}-\mu_{c}^{i}}{\sigma_{c}^{i}}+\beta_{c}^{i},
\end{equation}
where $\mu_{c}^{i}$ and $\sigma_{c}^{i}$ are the channel-wise mean and variance of the foreground feature in $i$ -th layer:
\begin{align}
&\mu_{c}^{i}=\frac{1}{\#\left\{M^{i}=1\right\}} \sum_{h, w} F_{h, w, c}^{i} \circ M_{h, w}^{i}\\
&\sigma_{c}^{i}=\sqrt{\frac{1}{\#\left\{M^{i}=1\right\}} \sum_{h, w}\left(F_{h, w, c}^{i} \circ M_{h, w}^{i}-\mu_{c}^{i}\right)^{2}+\epsilon}
\end{align}
The expression $\#\{x=k\}$ means the number of pixels equal to value $k$ in $x$. The $\gamma_{c}^{i}$ and $\beta_{c}^{i}$ are the mean and standard deviation of the activations of the background in channel $c$ of layer $i$ :
\begin{align}
&\gamma_{c}^{i}=\frac{1}{\#\left\{\bar{M}^{i}=1\right\}} \sum_{h, w} F_{h, w, c}^{i} \circ \bar{M}_{h, w}^{i}\\
&\beta_{c}^{i}=\sqrt{\frac{1}{\#\left\{\bar{M}^{i}=1\right\}} \sum_{h, w}\left(F_{h, w, c}^{i} \circ \bar{M}_{h, w}^{i}-\gamma_{c}^{i}\right)^{2}+\epsilon}
\end{align}
where $\bar{M}^{i}$ is the background mask in $i$ -th layer.
The effect of the experiment is shown in Figure \ref{harmony_pic}, and the results show that the foreground is significantly more consistent with the background.

\section{Experiments}
In this section, we report the evaluation results of our method in both synthetic and real-world datasets, where the test images are generated using foreground images with ground truth mattes and random background images.  We will introduce the implementation details firstly, and then the results on synthetic datasets and real-world datasets.
Finally, ablation experiments are carried out to verify the effect of each module and operation. 

\subsection{Implementation Details}
We use Pytorch \cite{pytorch} to implement our model and adopt ResNet-50 \cite{resnet} pretrained on ImageNet \cite{imagenet} as our backbone.
The training, validation, and testing splits of the dataset are all consistent with the previous methods \cite{modnet,human2_modnet,lfm}.

\textbf{Training setting:}
In the training stage, we resize each image to $512 \times 512$ with random horizontal flipping for the coarse module. In refine module, we crop the whole image to different patches instead of resizing the image.
We use Adam \cite{adam} with $\beta_{1}$ = 0.5 and $\beta_{2}$ = 0.999 as optimizer. 
For the coarse module, we use the warm-up and linear decay strategies with the maximum learning rate $5 \times 10^{-3}$ for the backbone and $0.05$ for other parts, and
the training process has a total of $10,000$ iterations.
For refine module, the learning rate is initialized to $1 \times 10^{-3}$ . 
The training lasts for $120, 000$ iterations with warm-up at the first $5, 000$ iterations and cosine learning rate decay \cite{mg_12,mg_28} for the rest.
The batch size of both is $16$.
The training of coarse module only uses binary labels, which means that alpha values greater than $0$ is set to $1$. 
In our real-world data set, we also only use the coarse binary label. 
The generating method of training and testing set is consistent with HAttMatting \cite{attention}.\\
\textbf{Test setting:}
During the inference stage, images are resized to $512 \times 512$ and then fed into the coarse module to obtain foreground without any other post-processing (e.g., CRF). 
The result of the generated coarse map is $1$-channel, which is copied to $3$-channel first, and then $6$-channel data is obtained through upsampling and concatenate with the original image, which is used as the input of refine module.\\
\textbf{Decoupled training strategy:}
\label{strategy}
We use the decoupled training strategy, which means the parameters of the coarse module are not updated when training the refine module.
% Previous work has done a lot of work on the weight of multi-losses.
% Admatting \cite{adamatting} employs learnable weights; 
% LFM \cite{lfm} and MOD \cite{human2_modnet} uses empirical hyper-parameters;
% Semantic \cite{semantic} adds various restrictions to different losses.
This strategy avoids balancing the loss of the coarse and refine modules, which are heavily dependent on data distribution and need to be recalibrated on different datasets.
Besides, decoupled strategy enables different modules to have their own more flexible learned hyper-parameters so that each module can fully optimize its own capabilities.\\
\textbf{Evaluation metrics:}
In line with previous methods, we use the official evaluation code \cite{dim}, and evaluate the results by Sum of Absolute Differences (SAD), Mean Squared Error (MSE), Gradient (Grad), and Connectivity (Conn) errors.

\subsection{Experimental Results}

\begin{figure}[htbp]
\centering
\includegraphics[scale=0.1600]{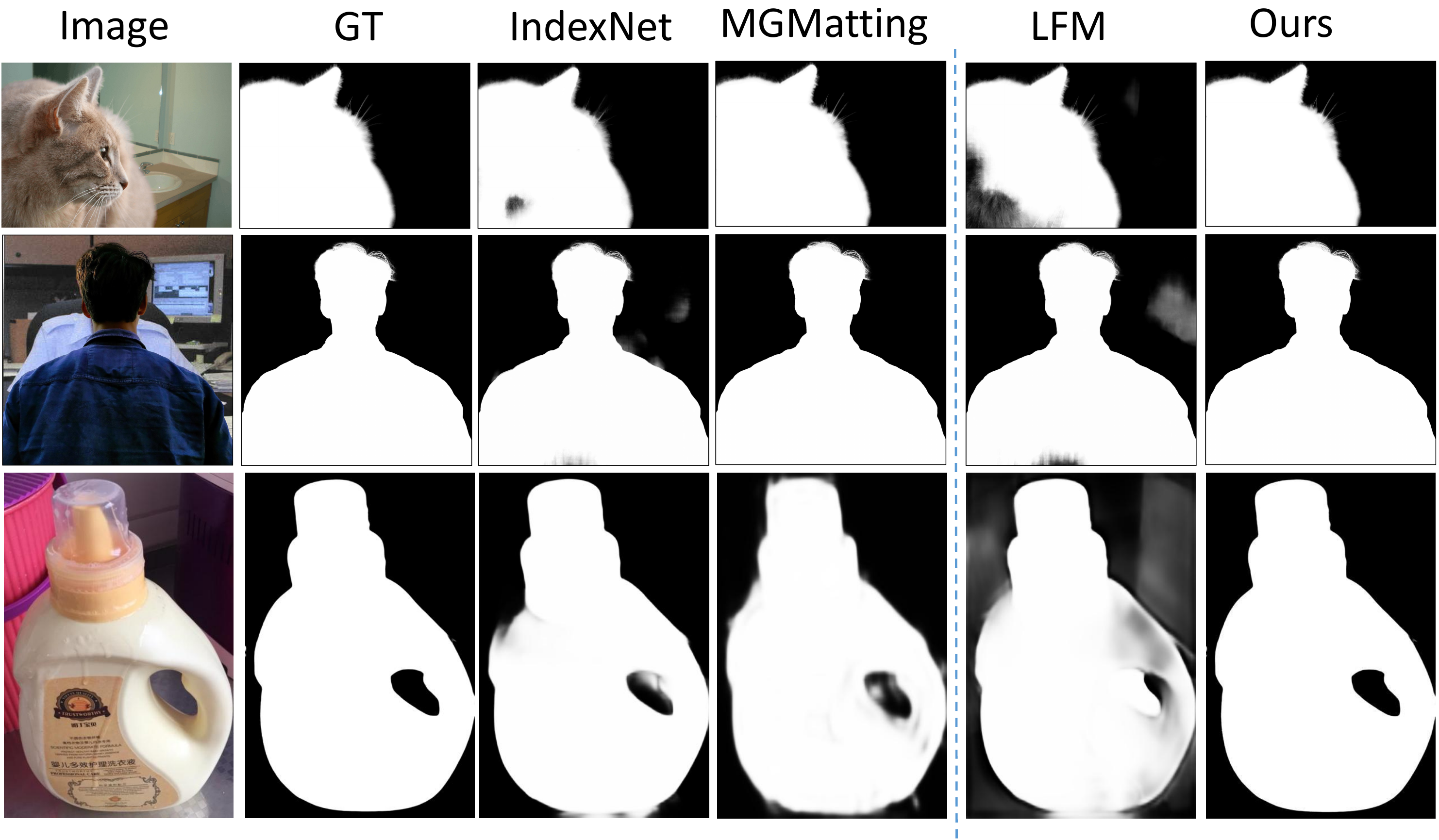}
\caption{
The left of the dotted line show results of trimap-based methods, and the right are those of trimap-free methods. The first, second, and third lines of images are from Composition-1k, Distinction-646, and our real world images, respectively. Both our approach and MGMatting are close to the true label in appearance. More results are in the supplementary material.
}
\label{pic_result}
\end{figure}

\begin{table}[htbp]
\centering
  \caption{Results on Composition-1k test set}
  \label{result_comp}
  \setlength{\tabcolsep}{0.4mm}{
  \scalebox{1}{
\begin{tabular}{cccccc}
\toprule
Trimap-&Method & SAD $\downarrow$ & MSE $\downarrow$ & Grad$\downarrow$ & Conn $\downarrow$\\
\midrule 
% Closed-Form  \cite{21_closedformmatting} & $168.1$ & 91 & $126.9$ & $167.9$ \\
% Shared  \cite{11_share_matting} & $125.3$ & $29$ & $144.2$ & $123.5$ \\
% Global \cite{15_globalmatting} & $156.8$ & $42$ & $112.2$ & $155.0$ \\
% KNN  \cite{6_knnmatting} & $175.4$ & 103 & $124.1$ & $176.4$ \\
% DCNN \cite{dcnn} & $115.8$ & $23$ & $107.3$ & $111.2$ \\

\multirow{9}{*}{Based}&Learning Based \cite{learning_based} & $113.9$ & 48 & $91.6$ & $122.2$ \\
&Information \cite{information_flow} & $70.3$ & $13$ & $42.8$ & $70.7$ \\
&DIM \cite{dim} & $50.4$ & 14 & $31.0$ & $50.8$ \\
&DCNN \cite{dcnn} & $115.8$ & $23$ & $107.3$ & $111.2$ \\
&AlphaGAN \cite{alphagan} & $90.94$ & $18$ & $93.92$ & $95.29$ \\
&IndexNet  \cite{indexnet} & $45.8$ & 13 & $25.9$ & $43.7$ \\
&AdaMatting \cite{adamatting} & $41.7$ & $10.2$ & $16.9$ & $34.4$ \\
&Context-Aware  \cite{18_context} & $35.8$ & $8.2$ & $17.3$ & $33.2$ \\
&GCA  \cite{24_gca} & $35.3$ & $9.1$ & $16.9$ & $32.5$ \\
&MGMatting \cite{mgmatting}  & $\textbf{32.1}$ & $\textbf{7.0}$ & $\textbf{14.0}$ & $\textbf{27.9}$ \\
\midrule 
\multirow{2}{*}{Free}&LFM \cite{lfm} & $58.4$ & $11.8$ & $41.6$ & $59.7$ \\
&MODNet \cite{modnet} & $47.1$ & $12.3$ & $37.0$ & $49.4$\\
&Ours & $\textbf{35.6}$ & $\textbf{8.7}$& $\textbf{17.6}$ & $\textbf{38.3}$ \\
\bottomrule
\end{tabular}
}
}
\end{table}

\textbf{Results on synthetic datasets.} 
The first two lines of Figure \ref{pic_result} show qualitative results and visual comparisons on synthetic datasets.
It can be observed that our results are better than the results of the LFM and comparable to MGMatting (SOTA of trimap-based matting).
The corresponding metrics are reported in Table \ref{result_comp}.
% 定量和定性的结果充分证明了使用显著性定义前景的合理，以及提出的coarse-to-fine网络结构的有效性。
Quantitative and qualitative results fully demonstrate the rationality of salient definition as the prospect and the effectiveness of the proposed coarse-to-fine network structure.
Moreover, our method perform better in generalization.
Distinction-646 is a recent synthetic matting benchmark dataset, which has greater diversity than Composition-1k.
As shown in Table \ref{result_dis}, our network trained on Composition-1k obtains SOTA on Distinction-646 dataset.
Adaptation to various scenes, the most critical flaw of the trimap-free approach, has made tremendous progress in our approach.

\begin{table}[htbp]
\centering
  \caption{Results on Distinction-646 test set.}
  \label{result_dis}
  \setlength{\tabcolsep}{0.4mm}{
  \begin{threeparttable}
  \scalebox{1}{
\begin{tabular}{cccccc}
\toprule
Trimap&Method & SAD $\downarrow$ & MSE $\downarrow$ & Grad $\downarrow$ & Conn $\downarrow$ \\
\midrule
\multirow{5}{*}{Based}&Learning Based  \cite{learning_based} & $105.1$ & 21 & $94.2$ & $110.4$ \\
&Information \cite{information_flow} & $78.9$ & $16$ & $58.7$ & $80.5$ \\
&DIM  \cite{dim} & $47.6$ & $9$ & $43.3$ & $55.9$ \\
&DCNN \cite{dcnn} & $103.8$ & $20$ & $82.5$ & $99.9$ \\
&HAttMatting \cite{attention} & $48.98$ & 9 & $41.57$ & $49.93$ \\
&Shared  \cite{11_share_matting} & $119.56$ & $26$ & $129.6$ & $114.3$ \\
&Global  \cite{15_globalmatting} & $135.56$ & $39$ & $119.5$ & $136.4$ \\
&Closed-Form  \cite{21_closedformmatting} & $105.73$ & 23 & $91.76$ & $114.5$ \\
&KNN  \cite{6_knnmatting} & $116.68$ & 25 & $103.1$ & $121.4$ \\
&MGMatting \cite{mgmatting}  & $\textbf{36.6}$ & $\textbf{7.2}$ & $\textbf{27.3}$ & $\textbf{35.1}$ \\
\midrule
\multirow{5}{*}{Free}&{LFM} \cite{lfm} & $44.6$ & $12.8$ & $47.6$ & $45.4$ \\
&MODNet \cite{modnet} & $41.7$ & $9.0$ & $35.9$ & $48.3$\\
&Ours & $\textbf{34.5}$ & $\textbf{7.8}$ & $\textbf{32.1}$ & $\textbf{42.6}$ \\

\cmidrule {2-6}
&LFM*  \cite{lfm} & $64.4$ & $21.8$ & $53.6$ & $69.4$ \\
&MODNet* \cite{modnet} & $70.2$ & $27.2$ & $57.9$ & $68.0$\\
&Ours* & $\textbf{44.7}$ & $\textbf{8.9}$ & $\textbf{42.6}$ & $\textbf{49.5}$ \\

\bottomrule
\end{tabular}
}
\begin{tablenotes} 
\small
		\item “*" represents trained on Composition-1k but tested on\\
		Distinction-646.
     \end{tablenotes} 
\end{threeparttable}
}
\end{table}

\begin{table}[htbp]
\centering
  \caption{Results on Real-19k test set.}
  \label{result_real}
  \setlength{\tabcolsep}{0.4mm}{
   \begin{threeparttable}
   \scalebox{1}{
\begin{tabular}{cccccc}
\toprule
Trimap&Methods & SAD $\downarrow$ & MSE $\downarrow$ & Grad $\downarrow$ & Conn $\downarrow$ \\
\midrule
% KNN \cite{6_knnmatting} & $116.68$ & 25 & $103.1$ & $121.4$ \\
\multirow{5}{*}{Based}&Deep Image   \cite{dim} & $52.5$ & $19$ & $46.1$ & $59.81$ \\
&Context-Aware  \cite{18_context} & $39.9$ & $14.2$ & $37.0$ & $43.1$ \\
&IndexNet  \cite{indexnet} & $44.7$ & $18.6$ & $23.6$ & $39.7$ \\
&MGMatting  \cite{mgmatting} & $42.8$ & $14.7$ & $\textbf{21.4}$ & $38.8$ \\
&Adamatting  \cite{adamatting} & $\textbf{38.1}$ & $\textbf{12.9}$ & $22.4$ & $\textbf{35.4}$ \\
\midrule

\multirow{6}{*}{Free}&{Deeplab} \cite{deeplab} & $44.6$ & $20.4$ & $37.8$ & $60.1$ \\
&{$U^{2}$Net} \cite{u2net} & $39.9$ & $14.6$ & $31.4$ & $39.5$ \\
&{LFM} \cite{lfm} & $58.4$ & $21.8$ & $41.6$ & $59.7$ \\
&MODNet \cite{modnet} & $44.4$ & $15.1$ & $42.3$ & $49.8$\\
&{Ours}  & $\textbf{35.4}$ & $\textbf{9.4}$ & $\textbf{19.6}$ &$\textbf{33.5}$ \\
\cmidrule {2-6}
&{LFM}* \cite{lfm} &$89.4$ & $31.6$ & $123.3$ & $115.7$ \\
&MODNet* \cite{modnet} & $92.4$ & $38.2$ & $157.0$ & $129.5$\\
&{Ours}*  & $\textbf{54.9}$ & $\textbf{17.1}$ & $\textbf{36.2}$  & $\textbf{58.4}$ \\
\bottomrule
\end{tabular}
}
\begin{tablenotes} 
\small
		\item “*" represents tested on real-world dataset finetuned	only with\\ coarse label.
     \end{tablenotes} 
\end{threeparttable}
}
\end{table}

\begin{figure}[htbp]
\centering
\includegraphics[scale=0.210]{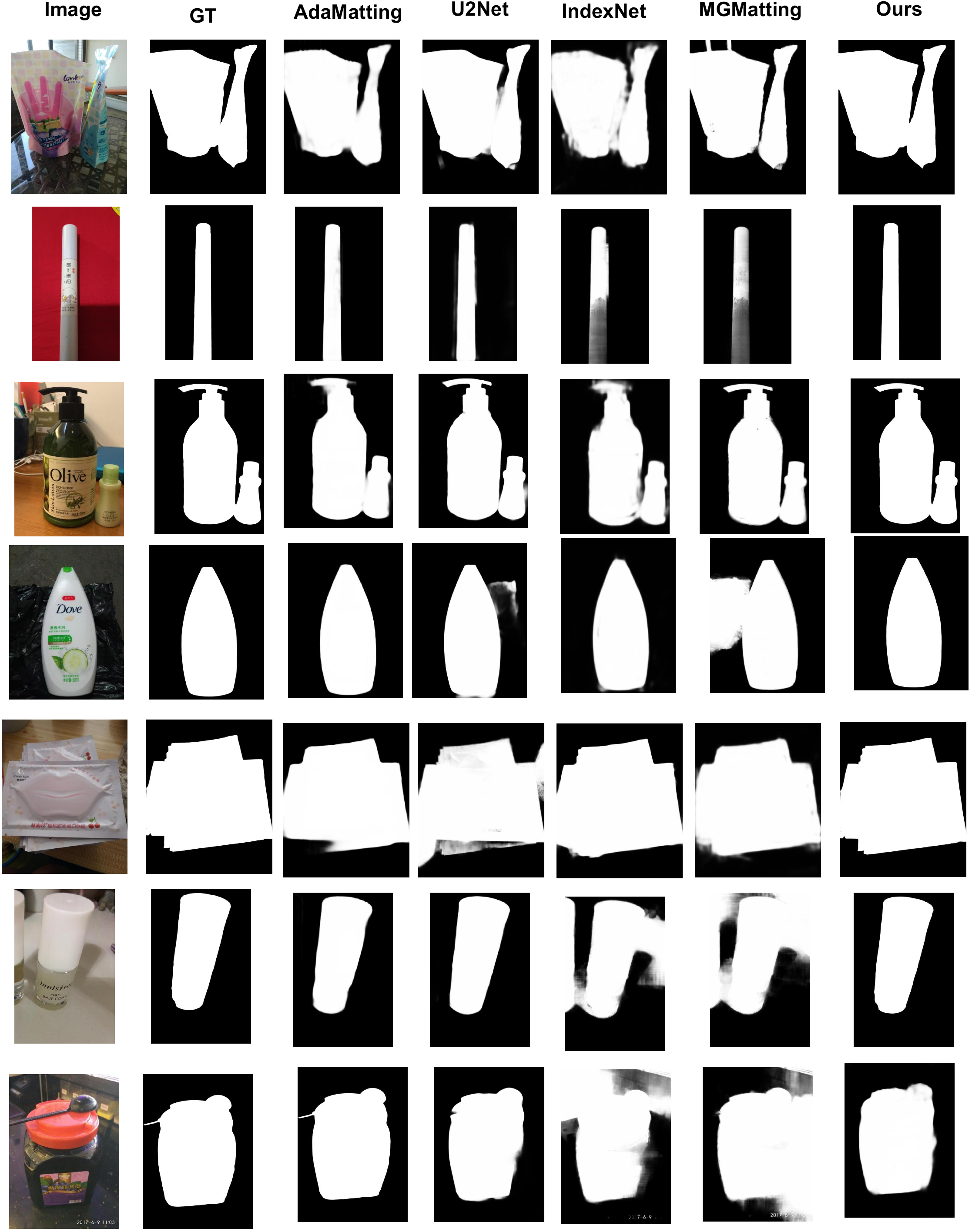}
\caption{
Comparison of results of different matting-methods Adamatting \cite{adamatting}, $U^{2}Net$ \cite{u2net}, IndexNet \cite{indexnet}, and MGMatting \cite{mgmatting} on  Real-19k.
}
\label{real_pic}
\end{figure}

\textbf{Results on real-world dataset.}
We display results of real images in Figure \ref{real_pic}, and our results are significantly better than others. The error of LFM is not only in the edge but also in the overall extraction of the foreground, including the missing foreground part and the interference in the background.
The corresponding metrics are reported in Table \ref{result_real}. 
As shown in Table \ref{result_real}, compared to the similar performance on synthetic datasets, the results of trimap-free approaches on real-world datasets are inferior to the trimap-based.
Methods marked with an asterisk means the network inherit the parameters trained with synthetic Composition-1k and just finetune the coarse module with the training set of Real-19k.
% Note that we only use the coarse label in Real-19k to train the coarse module, i.e., we do not need the alpha-level label of the real-world dataset but output alpha-level matte.
%The capacity of refining edge (unknown) regions trained on synthetic data can further apply to the real world.
As can be seen, previous works like LFM has worse generalization performance in real-world images than our MFC-Net.
\begin{figure*}[htbp]
\centering
\includegraphics[scale=0.12]{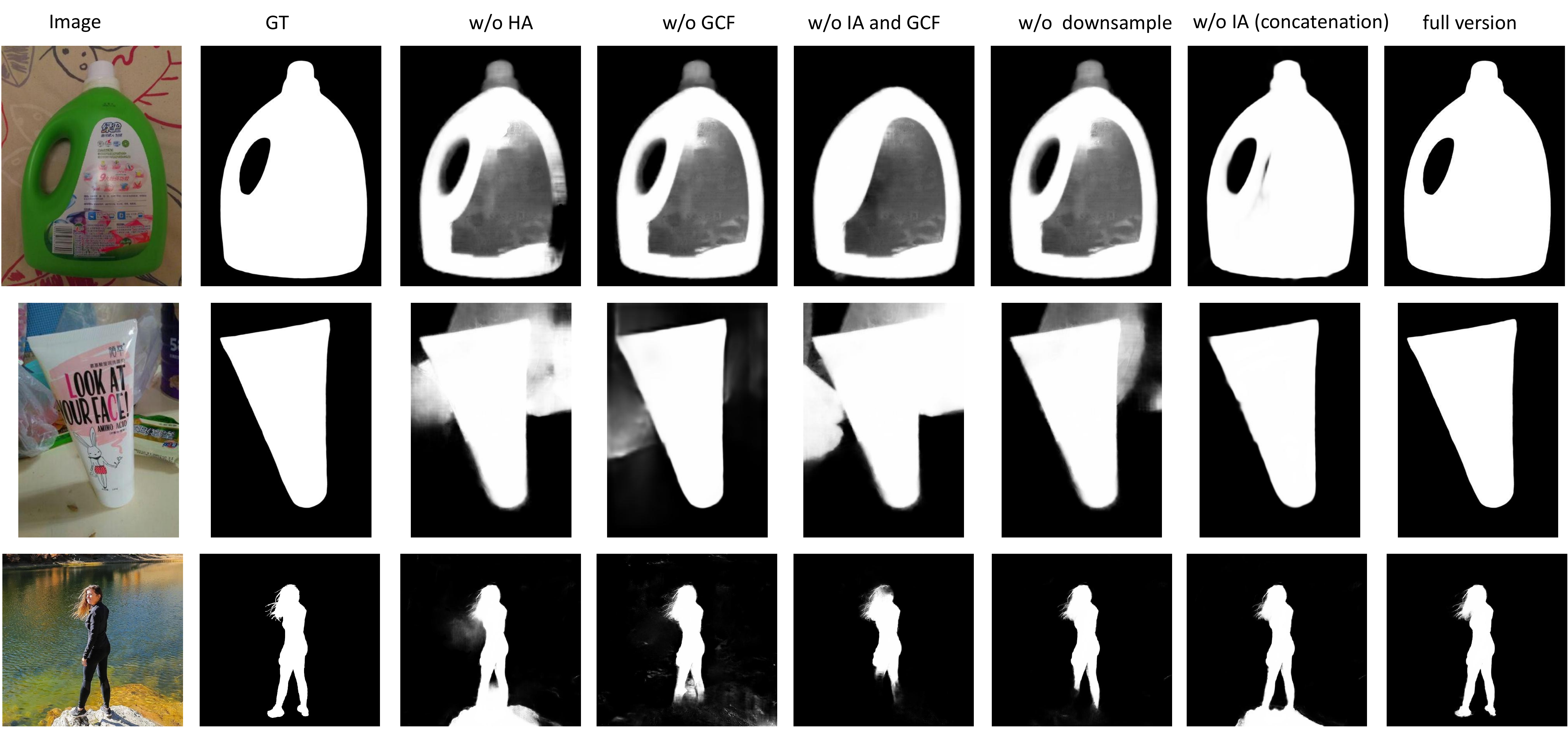}
\caption{
Visualized results of ablation experiments.
w/o IA stands means directly concatenating three kinds of information in channel dimension.
}
\label{pic_abla}
\end{figure*}

% \begin{table}[htbp]
% \centering
%   \caption{Ablation studies on Real-19k test set.}
%   \label{result_abla}
%     \scalebox{1}{
% \begin{tabular}{c|c|c|c|c|c|c}
% \toprule
% % \multirow{2}{*}{ Methods } & \multicolumn{2}{|c|} { Whole Image } & \multicolumn{2}{|c} {Unknown Area } \\
% % \cline{2-5} & SAD & MSE  & SAD & MSE \\
% HA &GCF&IA & SAD $\downarrow$ & MSE $\downarrow$ & Grad $\downarrow$ & Conn $\downarrow$ \\
% \midrule 
% &&& $77.3$ & $36.6$& $57.4$ & $72.2$ \\
% \Checkmark& && $63.9$ & $32.9$& $52.8$ & $67.5$ \\
% \Checkmark&\Checkmark&   & $41.3$ & $11.6$ & $28.9$ & $41.0$ \\
% \Checkmark&&\Checkmark   & $41.1$ & $16.2$& $35.9$ & $42.8$  \\
% &\Checkmark & \Checkmark  & $37.8$ & $12.7$& $24.3$ & $36.5$  \\
% \Checkmark  &\Checkmark &\Checkmark &$\textbf{35.4}$ & $\textbf{9.4}$& $\textbf{19.6}$ & $\textbf{33.5}$ \\

% \bottomrule
% \end{tabular}
% }
% \end{table}

\subsection{Ablation Studies}

\begin{table}[htbp]
\centering
  \caption{Ablation studies on Real-19k test set.}
  \label{result_abla_2}
    \begin{threeparttable}
    \scalebox{1}{
\begin{tabular}{c|c|c|c|c}
\toprule
% \multirow{2}{*}{ Methods } & \multicolumn{2}{|c|} { Whole Image } & \multicolumn{2}{|c} {Unknown Area } \\
% \cline{2-5} & SAD & MSE  & SAD & MSE \\
Methods & SAD $\downarrow$ & MSE $\downarrow$ & Grad $\downarrow$ & Conn $\downarrow$ \\
\midrule 
w/o HA & $37.8$ & $12.7$& $24.3$ & $36.5$ \\
w/o IA & $41.3$ & $11.6$ & $28.9$ & $41.0$\\
w/o GCF & $41.1$ & $16.2$& $35.9$ & $42.8$\\
w/o IA and GCF& $63.9$ & $32.9$& $52.8$ & $67.5$\\
\midrule 
% \midrule 
% w/o harmony (256) & $53.9$ & $30.6$ & $37.8$ & $37.5$\\
% w/ harmony (256) & $53.9$ & $30.6$ & $37.8$ & $37.5$\\
% \midrule 
w/o harmony & $43.6$ & $22.2$ & $30.3$ & $37.7$\\
w/ harmony &$\textbf{35.4}$ & $\textbf{9.4}$& $\textbf{19.6}$ & $\textbf{33.5}$ \\
% \midrule 
\midrule 
Original Size & $53.9$ & $30.6$ & $37.8$ & $37.5$ \\
$1/4$ Relative Size &$47.9$ & $29.3$ & $33.4$ & $39.3$\\
$1/2$ Relative Size& $59.7$ & $15.5$ & $41.1$ & $39.5$\\
256 Absolute Size & $89.4$ & $36.3$ & $59.8$ & $61.2$ \\
512 Absolute Size &$\textbf{35.4}$ & $\textbf{9.4}$& $\textbf{19.6}$ & $\textbf{33.5}$ \\
\bottomrule
\end{tabular}
}
\begin{tablenotes} 
\small
    \item w/o denotes without, and w/ denotes with.
     \end{tablenotes} 
\end{threeparttable}
\end{table}

% The effect of dwonsampling in coarse module and harmony in data generation is shown .\\
\textbf{The effect of components in coarse module:}
Table \ref{result_abla_2} shows the unique roles of each module.
Visual results are shown in Figure  \ref{pic_abla}.
HA is an initial enhancement of the foreground and a weakening of the background, which is indispensable.
MSE lost 3.3 when without HA compared with full-version.
After losing the GCF, images may not get terribly error, but the number of images that go wrong is huge.
The global feature provided by GCF can be considered as a norm feature.
The difference of foreground and background in various images is diverse, but the global features can normalize the difference to a unified range to facilitate the network learning of this difference.
No IA means concatenating the three kinds of information together on the channel dimension without any filter or mask. It can be seen that there is a 2.2 gap in MSE between direct concatenation and IA.
2.2 is also a considerable improvement over the already high accuracy.
With HA alone, the MSE is 32.9, but adding IA, it becomes 16.2, which also shows the importance of IA.
In brief, the three modules complement each other, strengthening the foreground and weakening the background, and losing either of them would result in a drop.

\textbf{Image harmony:}
Image harmony in synthetic image generation plays a crucial role in model accuracy and generalization. 
Without harmony, even training on the real world data set, the effect will decline.
MSE improves from 36.3 to 30.6 after harmony in downsampling of 256.
When cooperating with downsampling of 512, the addition of harmony improves MSE from 15.5 to 9.4. 
The significance of harmony shows in that it can increase the generalization and cross-scene capacity of the models. As shown in Table \ref{result_dis} and \ref{result_real}, the accuracy loss of our model is much less than that of LFM when converting to different scenes.
Compared with the baseline, the full version model brings a very significant performance improvement, surpassing all previous matting methods.

\textbf{Input size of coarse module:}
For the effect of downsampling, as can be seen from the second to the fifth line in Table \ref{result_abla_2}, even if all components are used but without downsampling, the effect is also poor. 
Base+origin means the original resolution is kept during both training and testing.
$\frac{1}{4}$ and $\frac{1}{2}$ mean downsampling to $\frac{1}{4}$ and $\frac{1}{2}$ of the original length, and 512 means downsampling to $512 \times 512$.
If the original resolution is maintained, crop operation has to be required during training, which leads to the convolution only learning local features, and the local features without surrounding dependent information can not accurately classify each pixel. 
However, it is also inappropriate to be excessive downsampling. After $\frac{1}{4}$ downsampling of the original image, the image size is only about $200 \times 200$, which will lose a lot of edge information and is fatal to the following refine module.
In short, the downsampling ratio is a trade-off choice, where downsampling makes the network extracting the full global features but losing edge details, and vice versa. All in all, 512 is a good choice.

\section{Conclusion}
This paper proposes targeted solutions to the current problems faced by trimap-free matting and achieves SOTA results on both synthetic and real-world images. Saliency is introduced to clearly define the foreground, and we design three modules to make full use of multiple features and extract accurate and complete foreground. We also establish a large-scale general matting dataset in the real world. 
In addition, our method owns the highly generalized capacity via image harmony in data generation, which is of great significance for practical use.
% \newpage
\bibliographystyle{ACM-Reference-Format}
\bibliography{main}


\begin{thebibliography}{92}

%%% the Web address.

\ifx \showCODEN    \undefined \def \showCODEN     #1{\unskip}     \fi
\ifx \showDOI      \undefined \def \showDOI       #1{#1}\fi
\ifx \showISBNx    \undefined \def \showISBNx     #1{\unskip}     \fi
\ifx \showISBNxiii \undefined \def \showISBNxiii  #1{\unskip}     \fi
\ifx \showISSN     \undefined \def \showISSN      #1{\unskip}     \fi
\ifx \showLCCN     \undefined \def \showLCCN      #1{\unskip}     \fi
\ifx \shownote     \undefined \def \shownote      #1{#1}          \fi
\ifx \showarticletitle \undefined \def \showarticletitle #1{#1}   \fi
\ifx \showURL      \undefined \def \showURL       {\relax}        \fi
% invisible to TeX
\providecommand\bibfield[2]{#2}
\providecommand\bibinfo[2]{#2}
\providecommand\natexlab[1]{#1}
\providecommand\showeprint[2][]{arXiv:#2}

\bibitem[Aksoy(2017)]%
        {information_flow}
\bibfield{author}{\bibinfo{person}{Ozan Aydin-Tunc Pollefeys~Marc Aksoy,
  Yagiz}.} \bibinfo{year}{2017}\natexlab{}.
\newblock \showarticletitle{Designing effective inter-pixel information flow
  for natural image matting}. In \bibinfo{booktitle}{\emph{Proceedings of the
  IEEE Conference on Computer Vision and Pattern Recognition}}.
  \bibinfo{pages}{29--37}.
\newblock


\bibitem[Bai et~al\mbox{.}(2019)]%
        {2}
\bibfield{author}{\bibinfo{person}{Yutong Bai}, \bibinfo{person}{Qing Liu},
  \bibinfo{person}{Lingxi Xie}, \bibinfo{person}{Weichao Qiu},
  \bibinfo{person}{Yan Zheng}, {and} \bibinfo{person}{Alan~L Yuille}.}
  \bibinfo{year}{2019}\natexlab{}.
\newblock \showarticletitle{Semantic part detection via matching: Learning to
  generalize to novel viewpoints from limited training data}. In
  \bibinfo{booktitle}{\emph{Proceedings of the IEEE/CVF International
  Conference on Computer Vision}}. \bibinfo{pages}{7535--7545}.
\newblock


\bibitem[Bai et~al\mbox{.}(2020)]%
        {3}
\bibfield{author}{\bibinfo{person}{Yutong Bai}, \bibinfo{person}{Angtian Wang},
  \bibinfo{person}{Adam Kortylewski}, {and} \bibinfo{person}{Alan Yuille}.}
  \bibinfo{year}{2020}\natexlab{}.
\newblock \showarticletitle{CoKe: Localized Contrastive Learning for Robust
  Keypoint Detection}.
\newblock \bibinfo{journal}{\emph{arXiv preprint arXiv:2009.14115}}
  (\bibinfo{year}{2020}).
\newblock


\bibitem[Bouville(2008)]%
        {c:22}
\bibfield{author}{\bibinfo{person}{Mathieu Bouville}.}
  \bibinfo{year}{2008}\natexlab{}.
\newblock \bibinfo{title}{Crime and punishment in scientific research}.
\newblock
\newblock
\showeprint[arxiv]{0803.4058}~[physics.soc-ph]


\bibitem[Cai et~al\mbox{.}(2019)]%
        {adamatting}
\bibfield{author}{\bibinfo{person}{Shaofan Cai}, \bibinfo{person}{Xiaoshuai
  Zhang}, \bibinfo{person}{Haoqiang Fan}, \bibinfo{person}{Haibin Huang},
  \bibinfo{person}{Jiangyu Liu}, \bibinfo{person}{Jiaming Liu},
  \bibinfo{person}{Jiaying Liu}, \bibinfo{person}{Jue Wang}, {and}
  \bibinfo{person}{Jian Sun}.} \bibinfo{year}{2019}\natexlab{}.
\newblock \showarticletitle{Disentangled image matting}. In
  \bibinfo{booktitle}{\emph{Proceedings of the IEEE/CVF International
  Conference on Computer Vision}}. \bibinfo{pages}{8819--8828}.
\newblock


\bibitem[Chen and Kae(2019)]%
        {harmony_1}
\bibfield{author}{\bibinfo{person}{Bor-Chun Chen} {and} \bibinfo{person}{Andrew
  Kae}.} \bibinfo{year}{2019}\natexlab{}.
\newblock \showarticletitle{Toward realistic image compositing with adversarial
  learning}. In \bibinfo{booktitle}{\emph{Proceedings of the IEEE/CVF
  Conference on Computer Vision and Pattern Recognition}}.
  \bibinfo{pages}{8415--8424}.
\newblock


\bibitem[Chen(2018)]%
        {deeplab}
\bibfield{author}{\bibinfo{person}{Liang-Chieh Chen}.}
  \bibinfo{year}{2018}\natexlab{}.
\newblock \showarticletitle{Encoder-decoder with atrous separable convolution
  for semantic image segmentation}. In \bibinfo{booktitle}{\emph{Proceedings of
  the European conference on computer vision (ECCV)}}.
  \bibinfo{pages}{801--818}.
\newblock


\bibitem[Chen et~al\mbox{.}(2017)]%
        {5_deeplab}
\bibfield{author}{\bibinfo{person}{Liang-Chieh Chen}, \bibinfo{person}{George
  Papandreou}, \bibinfo{person}{Iasonas Kokkinos}, \bibinfo{person}{Kevin
  Murphy}, {and} \bibinfo{person}{Alan~L Yuille}.}
  \bibinfo{year}{2017}\natexlab{}.
\newblock \showarticletitle{Deeplab: Semantic image segmentation with deep
  convolutional nets, atrous convolution, and fully connected crfs}.
\newblock \bibinfo{journal}{\emph{IEEE transactions on pattern analysis and
  machine intelligence}} \bibinfo{volume}{40}, \bibinfo{number}{4}
  (\bibinfo{year}{2017}), \bibinfo{pages}{834--848}.
\newblock


\bibitem[Chen et~al\mbox{.}(2018)]%
        {human1}
\bibfield{author}{\bibinfo{person}{Quan Chen}, \bibinfo{person}{Tiezheng Ge},
  \bibinfo{person}{Yanyu Xu}, \bibinfo{person}{Zhiqiang Zhang},
  \bibinfo{person}{Xinxin Yang}, {and} \bibinfo{person}{Kun Gai}.}
  \bibinfo{year}{2018}\natexlab{}.
\newblock \showarticletitle{Semantic human matting}. In
  \bibinfo{booktitle}{\emph{Proceedings of the 26th ACM international
  conference on Multimedia}}. \bibinfo{pages}{618--626}.
\newblock


\bibitem[Chen et~al\mbox{.}(2013)]%
        {6_knnmatting}
\bibfield{author}{\bibinfo{person}{Qifeng Chen}, \bibinfo{person}{Dingzeyu Li},
  {and} \bibinfo{person}{Chi-Keung Tang}.} \bibinfo{year}{2013}\natexlab{}.
\newblock \showarticletitle{KNN matting}.
\newblock \bibinfo{journal}{\emph{IEEE transactions on pattern analysis and
  machine intelligence}} \bibinfo{volume}{35}, \bibinfo{number}{9}
  (\bibinfo{year}{2013}), \bibinfo{pages}{2175--2188}.
\newblock


\bibitem[Chen et~al\mbox{.}(2019)]%
        {glnet}
\bibfield{author}{\bibinfo{person}{Wuyang Chen}, \bibinfo{person}{Ziyu Jiang},
  \bibinfo{person}{Zhangyang Wang}, \bibinfo{person}{Kexin Cui}, {and}
  \bibinfo{person}{Xiaoning Qian}.} \bibinfo{year}{2019}\natexlab{}.
\newblock \showarticletitle{Collaborative global-local networks for
  memory-efficient segmentation of ultra-high resolution images}. In
  \bibinfo{booktitle}{\emph{Proceedings of the IEEE/CVF Conference on Computer
  Vision and Pattern Recognition}}. \bibinfo{pages}{8924--8933}.
\newblock


\bibitem[Chen et~al\mbox{.}(2020)]%
        {gcpanet}
\bibfield{author}{\bibinfo{person}{Zuyao Chen}, \bibinfo{person}{Qianqian Xu},
  \bibinfo{person}{Runmin Cong}, {and} \bibinfo{person}{Qingming Huang}.}
  \bibinfo{year}{2020}\natexlab{}.
\newblock \showarticletitle{Global context-aware progressive aggregation
  network for salient object detection}. In
  \bibinfo{booktitle}{\emph{Proceedings of the AAAI Conference on Artificial
  Intelligence}}, Vol.~\bibinfo{volume}{34}. \bibinfo{pages}{10599--10606}.
\newblock


\bibitem[Cho(2016)]%
        {dcnn}
\bibfield{author}{\bibinfo{person}{Tai Yu-Wing Kweon~Inso Cho, Donghyeon}.}
  \bibinfo{year}{2016}\natexlab{}.
\newblock \showarticletitle{Natural image matting using deep convolutional
  neural networks}. In \bibinfo{booktitle}{\emph{European Conference on
  Computer Vision}}. Springer, \bibinfo{pages}{626--643}.
\newblock


\bibitem[Chuang et~al\mbox{.}(2001)]%
        {7}
\bibfield{author}{\bibinfo{person}{Yung-Yu Chuang}, \bibinfo{person}{Brian
  Curless}, \bibinfo{person}{David~H Salesin}, {and} \bibinfo{person}{Richard
  Szeliski}.} \bibinfo{year}{2001}\natexlab{}.
\newblock \showarticletitle{A bayesian approach to digital matting}. In
  \bibinfo{booktitle}{\emph{Proceedings of the 2001 IEEE Computer Society
  Conference on Computer Vision and Pattern Recognition. CVPR 2001}},
  Vol.~\bibinfo{volume}{2}. IEEE, \bibinfo{pages}{II--II}.
\newblock


\bibitem[Clancey(1979)]%
        {c:79}
\bibfield{author}{\bibinfo{person}{William~J. Clancey}.}
  \bibinfo{year}{1979}\natexlab{}.
\newblock \emph{\bibinfo{title}{{Transfer of Rule-Based Expertise through a
  Tutorial Dialogue}}}.
\newblock {Ph.D.} diss. \bibinfo{school}{Dept.\ of Computer Science, Stanford
  Univ.}, \bibinfo{address}{Stanford, Calif.}
\newblock


\bibitem[Clancey(1983)]%
        {c:83}
\bibfield{author}{\bibinfo{person}{William~J. Clancey}.}
  \bibinfo{year}{1983}\natexlab{}.
\newblock \showarticletitle{{Communication, Simulation, and Intelligent Agents:
  Implications of Personal Intelligent Machines for Medical Education}}. In
  \bibinfo{booktitle}{\emph{Proceedings of the Eighth International Joint
  Conference on Artificial Intelligence {(IJCAI-83)}}}.
  \bibinfo{publisher}{{IJCAI Organization}}, \bibinfo{address}{Menlo Park,
  Calif}, \bibinfo{pages}{556--560}.
\newblock


\bibitem[Clancey(1984)]%
        {c:84}
\bibfield{author}{\bibinfo{person}{William~J. Clancey}.}
  \bibinfo{year}{1984}\natexlab{}.
\newblock \showarticletitle{{Classification Problem Solving}}. In
  \bibinfo{booktitle}{\emph{Proceedings of the Fourth National Conference on
  Artificial Intelligence}}. \bibinfo{publisher}{AAAI Press},
  \bibinfo{address}{Menlo Park, Calif.}, \bibinfo{pages}{45--54}.
\newblock


\bibitem[Clancey(2021)]%
        {c:21}
\bibfield{author}{\bibinfo{person}{William~J. Clancey}.}
  \bibinfo{year}{2021}\natexlab{}.
\newblock \bibinfo{title}{{The Engineering of Qualitative Models}}.
  (\bibinfo{year}{2021}).
\newblock
\newblock
\shownote{Forthcoming}.


\bibitem[Deng et~al\mbox{.}(2009)]%
        {imagenet}
\bibfield{author}{\bibinfo{person}{Jia Deng}, \bibinfo{person}{Wei Dong},
  \bibinfo{person}{Richard Socher}, \bibinfo{person}{Li-Jia Li},
  \bibinfo{person}{Kai Li}, {and} \bibinfo{person}{Li Fei-Fei}.}
  \bibinfo{year}{2009}\natexlab{}.
\newblock \showarticletitle{Imagenet: A large-scale hierarchical image
  database}. In \bibinfo{booktitle}{\emph{2009 IEEE conference on computer
  vision and pattern recognition}}. Ieee, \bibinfo{pages}{248--255}.
\newblock


\bibitem[Engelmore and Morgan(1986)]%
        {em:86}
\bibfield{editor}{\bibinfo{person}{Robert Engelmore} {and}
  \bibinfo{person}{Anthony Morgan}} (Eds.). \bibinfo{year}{1986}\natexlab{}.
\newblock \bibinfo{booktitle}{\emph{Blackboard Systems}}.
\newblock \bibinfo{publisher}{Addison-Wesley}, \bibinfo{address}{Reading,
  Mass.}
\newblock


\bibitem[Everingham and Winn(2011)]%
        {pascal_voc}
\bibfield{author}{\bibinfo{person}{Mark Everingham} {and} \bibinfo{person}{John
  Winn}.} \bibinfo{year}{2011}\natexlab{}.
\newblock \showarticletitle{The pascal visual object classes challenge 2012
  (voc2012) development kit}.
\newblock \bibinfo{journal}{\emph{Pattern Analysis, Statistical Modelling and
  Computational Learning, Tech. Rep}}  \bibinfo{volume}{8}
  (\bibinfo{year}{2011}), \bibinfo{pages}{5}.
\newblock


\bibitem[Gastal and Oliveira(2010)]%
        {11}
\bibfield{author}{\bibinfo{person}{Eduardo~SL Gastal} {and}
  \bibinfo{person}{Manuel~M Oliveira}.} \bibinfo{year}{2010}\natexlab{}.
\newblock \showarticletitle{Shared sampling for real-time alpha matting}. In
  \bibinfo{booktitle}{\emph{Computer Graphics Forum}},
  Vol.~\bibinfo{volume}{29}. Wiley Online Library, \bibinfo{pages}{575--584}.
\newblock


\bibitem[Gastal(2010)]%
        {11_share_matting}
\bibfield{author}{\bibinfo{person}{Oliveira-Manuel~M Gastal, Eduardo~SL}.}
  \bibinfo{year}{2010}\natexlab{}.
\newblock \showarticletitle{Shared sampling for real-time alpha matting}. In
  \bibinfo{booktitle}{\emph{Computer Graphics Forum}},
  Vol.~\bibinfo{volume}{29}. Wiley Online Library, \bibinfo{pages}{575--584}.
\newblock


\bibitem[Goyal et~al\mbox{.}(2017)]%
        {mg_12}
\bibfield{author}{\bibinfo{person}{Priya Goyal}, \bibinfo{person}{Piotr
  Doll{\'a}r}, \bibinfo{person}{Ross Girshick}, \bibinfo{person}{Pieter
  Noordhuis}, \bibinfo{person}{Lukasz Wesolowski}, \bibinfo{person}{Aapo
  Kyrola}, \bibinfo{person}{Andrew Tulloch}, \bibinfo{person}{Yangqing Jia},
  {and} \bibinfo{person}{Kaiming He}.} \bibinfo{year}{2017}\natexlab{}.
\newblock \showarticletitle{Accurate, large minibatch sgd: Training imagenet in
  1 hour}.
\newblock \bibinfo{journal}{\emph{arXiv preprint arXiv:1706.02677}}
  (\bibinfo{year}{2017}).
\newblock


\bibitem[Hasling et~al\mbox{.}(1984)]%
        {hcr:83}
\bibfield{author}{\bibinfo{person}{Diane~Warner Hasling},
  \bibinfo{person}{William~J. Clancey}, {and} \bibinfo{person}{Glenn Rennels}.}
  \bibinfo{year}{1984}\natexlab{}.
\newblock \showarticletitle{Strategic explanations for a diagnostic
  consultation system}.
\newblock \bibinfo{journal}{\emph{International Journal of Man-Machine
  Studies}} \bibinfo{volume}{20}, \bibinfo{number}{1} (\bibinfo{year}{1984}),
  \bibinfo{pages}{3--19}.
\newblock
\showISSN{0020-7373}
\urldef\tempurl%
\url{https://doi.org/10.1016/S0020-7373(84)80003-6}
\showDOI{\tempurl}


\bibitem[Hasling et~al\mbox{.}(1983)]%
        {hcrt:83}
\bibfield{author}{\bibinfo{person}{Diane~Warner Hasling},
  \bibinfo{person}{William~J. Clancey}, \bibinfo{person}{Glenn~R. Rennels},
  {and} \bibinfo{person}{Thomas Test}.} \bibinfo{year}{1983}\natexlab{}.
\newblock \showarticletitle{{Strategic Explanations in
  Consultation---Duplicate}}.
\newblock \bibinfo{journal}{\emph{The International Journal of Man-Machine
  Studies}} \bibinfo{volume}{20}, \bibinfo{number}{1} (\bibinfo{year}{1983}),
  \bibinfo{pages}{3--19}.
\newblock


\bibitem[He et~al\mbox{.}(2017)]%
        {maskrcnn}
\bibfield{author}{\bibinfo{person}{Kaiming He}, \bibinfo{person}{Georgia
  Gkioxari}, \bibinfo{person}{Piotr Doll{\'a}r}, {and} \bibinfo{person}{Ross
  Girshick}.} \bibinfo{year}{2017}\natexlab{}.
\newblock \showarticletitle{Mask r-cnn}. In
  \bibinfo{booktitle}{\emph{Proceedings of the IEEE international conference on
  computer vision}}. \bibinfo{pages}{2961--2969}.
\newblock


\bibitem[He et~al\mbox{.}(2011)]%
        {15_globalmatting}
\bibfield{author}{\bibinfo{person}{Kaiming He}, \bibinfo{person}{Christoph
  Rhemann}, \bibinfo{person}{Carsten Rother}, \bibinfo{person}{Xiaoou Tang},
  {and} \bibinfo{person}{Jian Sun}.} \bibinfo{year}{2011}\natexlab{}.
\newblock \showarticletitle{A global sampling method for alpha matting}. In
  \bibinfo{booktitle}{\emph{CVPR 2011}}. IEEE, \bibinfo{pages}{2049--2056}.
\newblock


\bibitem[He et~al\mbox{.}(2010)]%
        {16}
\bibfield{author}{\bibinfo{person}{Kaiming He}, \bibinfo{person}{Jian Sun},
  {and} \bibinfo{person}{Xiaoou Tang}.} \bibinfo{year}{2010}\natexlab{}.
\newblock \showarticletitle{Fast matting using large kernel matting laplacian
  matrices}. In \bibinfo{booktitle}{\emph{2010 IEEE Computer Society Conference
  on Computer Vision and Pattern Recognition}}. IEEE,
  \bibinfo{pages}{2165--2172}.
\newblock


\bibitem[He et~al\mbox{.}(2016)]%
        {resnet}
\bibfield{author}{\bibinfo{person}{Kaiming He}, \bibinfo{person}{Xiangyu
  Zhang}, \bibinfo{person}{Shaoqing Ren}, {and} \bibinfo{person}{Jian Sun}.}
  \bibinfo{year}{2016}\natexlab{}.
\newblock \showarticletitle{Deep residual learning for image recognition}. In
  \bibinfo{booktitle}{\emph{Proceedings of the IEEE conference on computer
  vision and pattern recognition}}. \bibinfo{pages}{770--778}.
\newblock


\bibitem[Hou and Liu(2019)]%
        {18_context}
\bibfield{author}{\bibinfo{person}{Qiqi Hou} {and} \bibinfo{person}{Feng Liu}.}
  \bibinfo{year}{2019}\natexlab{}.
\newblock \showarticletitle{Context-aware image matting for simultaneous
  foreground and alpha estimation}. In \bibinfo{booktitle}{\emph{Proceedings of
  the IEEE/CVF International Conference on Computer Vision}}.
  \bibinfo{pages}{4130--4139}.
\newblock


\bibitem[Hu and Clark(2019)]%
        {instance}
\bibfield{author}{\bibinfo{person}{Guanqing Hu} {and} \bibinfo{person}{James
  Clark}.} \bibinfo{year}{2019}\natexlab{}.
\newblock \showarticletitle{Instance segmentation based semantic matting for
  compositing applications}. In \bibinfo{booktitle}{\emph{2019 16th Conference
  on Computer and Robot Vision (CRV)}}. IEEE, \bibinfo{pages}{135--142}.
\newblock


\bibitem[Ke et~al\mbox{.}(2020)]%
        {human2_modnet}
\bibfield{author}{\bibinfo{person}{Zhanghan Ke}, \bibinfo{person}{Kaican Li},
  \bibinfo{person}{Yurou Zhou}, \bibinfo{person}{Qiuhua Wu},
  \bibinfo{person}{Xiangyu Mao}, \bibinfo{person}{Qiong Yan}, {and}
  \bibinfo{person}{Rynson~WH Lau}.} \bibinfo{year}{2020}\natexlab{}.
\newblock \showarticletitle{Is a Green Screen Really Necessary for Real-Time
  Human Matting?}
\newblock \bibinfo{journal}{\emph{arXiv preprint arXiv:2011.11961}}
  (\bibinfo{year}{2020}).
\newblock


\bibitem[Ke et~al\mbox{.}(2022)]%
        {modnet}
\bibfield{author}{\bibinfo{person}{Zhanghan Ke}, \bibinfo{person}{Jiayu Sun},
  \bibinfo{person}{Kaican Li}, \bibinfo{person}{Qiong Yan}, {and}
  \bibinfo{person}{Rynson~WH Lau}.} \bibinfo{year}{2022}\natexlab{}.
\newblock \showarticletitle{Modnet: Real-time trimap-free portrait matting via
  objective decomposition}. In \bibinfo{booktitle}{\emph{Proceedings of the
  AAAI Conference on Artificial Intelligence}}, Vol.~\bibinfo{volume}{36}.
  \bibinfo{pages}{1140--1147}.
\newblock


\bibitem[Lee and Wu(2011)]%
        {20}
\bibfield{author}{\bibinfo{person}{Philip Lee} {and} \bibinfo{person}{Ying
  Wu}.} \bibinfo{year}{2011}\natexlab{}.
\newblock \showarticletitle{Nonlocal matting}. In
  \bibinfo{booktitle}{\emph{CVPR 2011}}. IEEE, \bibinfo{pages}{2193--2200}.
\newblock


\bibitem[Levin et~al\mbox{.}(2007)]%
        {21_closedformmatting}
\bibfield{author}{\bibinfo{person}{Anat Levin}, \bibinfo{person}{Dani
  Lischinski}, {and} \bibinfo{person}{Yair Weiss}.}
  \bibinfo{year}{2007}\natexlab{}.
\newblock \showarticletitle{A closed-form solution to natural image matting}.
\newblock \bibinfo{journal}{\emph{IEEE transactions on pattern analysis and
  machine intelligence}} \bibinfo{volume}{30}, \bibinfo{number}{2}
  (\bibinfo{year}{2007}), \bibinfo{pages}{228--242}.
\newblock


\bibitem[Levin et~al\mbox{.}(2008)]%
        {22}
\bibfield{author}{\bibinfo{person}{Anat Levin}, \bibinfo{person}{Alex
  Rav-Acha}, {and} \bibinfo{person}{Dani Lischinski}.}
  \bibinfo{year}{2008}\natexlab{}.
\newblock \showarticletitle{Spectral matting}.
\newblock \bibinfo{journal}{\emph{IEEE transactions on pattern analysis and
  machine intelligence}} \bibinfo{volume}{30}, \bibinfo{number}{10}
  (\bibinfo{year}{2008}), \bibinfo{pages}{1699--1712}.
\newblock


\bibitem[Li et~al\mbox{.}(2020)]%
        {animal}
\bibfield{author}{\bibinfo{person}{Jizhizi Li}, \bibinfo{person}{Jing Zhang},
  \bibinfo{person}{Stephen~J Maybank}, {and} \bibinfo{person}{Dacheng Tao}.}
  \bibinfo{year}{2020}\natexlab{}.
\newblock \showarticletitle{End-to-end Animal Image Matting}.
\newblock \bibinfo{journal}{\emph{arXiv preprint arXiv:2010.16188}}
  (\bibinfo{year}{2020}).
\newblock


\bibitem[Li et~al\mbox{.}(2021)]%
        {aim}
\bibfield{author}{\bibinfo{person}{Jizhizi Li}, \bibinfo{person}{Jing Zhang},
  {and} \bibinfo{person}{Dacheng Tao}.} \bibinfo{year}{2021}\natexlab{}.
\newblock \showarticletitle{Deep automatic natural image matting}.
\newblock \bibinfo{journal}{\emph{arXiv preprint arXiv:2107.07235}}
  (\bibinfo{year}{2021}).
\newblock


\bibitem[Li and Lu(2020)]%
        {24_gca}
\bibfield{author}{\bibinfo{person}{Yaoyi Li} {and} \bibinfo{person}{Hongtao
  Lu}.} \bibinfo{year}{2020}\natexlab{}.
\newblock \showarticletitle{Natural image matting via guided contextual
  attention}. In \bibinfo{booktitle}{\emph{Proceedings of the AAAI Conference
  on Artificial Intelligence}}, Vol.~\bibinfo{volume}{34}.
  \bibinfo{pages}{11450--11457}.
\newblock


\bibitem[Lin et~al\mbox{.}(2013)]%
        {gap_chen}
\bibfield{author}{\bibinfo{person}{Min Lin}, \bibinfo{person}{Qiang Chen},
  {and} \bibinfo{person}{Shuicheng Yan}.} \bibinfo{year}{2013}\natexlab{}.
\newblock \showarticletitle{Network in network}.
\newblock \bibinfo{journal}{\emph{arXiv preprint arXiv:1312.4400}}
  (\bibinfo{year}{2013}).
\newblock


\bibitem[Lin et~al\mbox{.}(2020)]%
        {backv2}
\bibfield{author}{\bibinfo{person}{Shanchuan Lin}, \bibinfo{person}{Andrey
  Ryabtsev}, \bibinfo{person}{Soumyadip Sengupta}, \bibinfo{person}{Brian
  Curless}, \bibinfo{person}{Steve Seitz}, {and} \bibinfo{person}{Ira
  Kemelmacher-Shlizerman}.} \bibinfo{year}{2020}\natexlab{}.
\newblock \showarticletitle{Real-Time High-Resolution Background Matting}.
\newblock \bibinfo{journal}{\emph{arXiv preprint arXiv:2012.07810}}
  (\bibinfo{year}{2020}).
\newblock


\bibitem[Ling et~al\mbox{.}(2021)]%
        {rainnet}
\bibfield{author}{\bibinfo{person}{Jun Ling}, \bibinfo{person}{Han Xue},
  \bibinfo{person}{Li Song}, \bibinfo{person}{Rong Xie}, {and}
  \bibinfo{person}{Xiao Gu}.} \bibinfo{year}{2021}\natexlab{}.
\newblock \showarticletitle{Region-aware Adaptive Instance Normalization for
  Image Harmonization}. In \bibinfo{booktitle}{\emph{Proceedings of the
  IEEE/CVF Conference on Computer Vision and Pattern Recognition}}.
  \bibinfo{pages}{9361--9370}.
\newblock


\bibitem[Liu et~al\mbox{.}(2020b)]%
        {f2net}
\bibfield{author}{\bibinfo{person}{Daizong Liu}, \bibinfo{person}{Dongdong Yu},
  \bibinfo{person}{Changhu Wang}, {and} \bibinfo{person}{Pan Zhou}.}
  \bibinfo{year}{2020}\natexlab{b}.
\newblock \showarticletitle{F2Net: Learning to Focus on the Foreground for
  Unsupervised Video Object Segmentation}.
\newblock \bibinfo{journal}{\emph{arXiv preprint arXiv:2012.02534}}
  (\bibinfo{year}{2020}).
\newblock


\bibitem[Liu et~al\mbox{.}(2020a)]%
        {boost}
\bibfield{author}{\bibinfo{person}{Jinlin Liu}, \bibinfo{person}{Yuan Yao},
  \bibinfo{person}{Wendi Hou}, \bibinfo{person}{Miaomiao Cui},
  \bibinfo{person}{Xuansong Xie}, \bibinfo{person}{Changshui Zhang}, {and}
  \bibinfo{person}{Xian-Sheng Hua}.} \bibinfo{year}{2020}\natexlab{a}.
\newblock \showarticletitle{Boosting semantic human matting with coarse
  annotations}. In \bibinfo{booktitle}{\emph{Proceedings of the IEEE/CVF
  Conference on Computer Vision and Pattern Recognition}}.
  \bibinfo{pages}{8563--8572}.
\newblock


\bibitem[Liu et~al\mbox{.}(2019)]%
        {poolnet}
\bibfield{author}{\bibinfo{person}{Jiang-Jiang Liu}, \bibinfo{person}{Qibin
  Hou}, \bibinfo{person}{Ming-Ming Cheng}, \bibinfo{person}{Jiashi Feng}, {and}
  \bibinfo{person}{Jianmin Jiang}.} \bibinfo{year}{2019}\natexlab{}.
\newblock \showarticletitle{A simple pooling-based design for real-time salient
  object detection}. In \bibinfo{booktitle}{\emph{Proceedings of the IEEE/CVF
  Conference on Computer Vision and Pattern Recognition}}.
  \bibinfo{pages}{3917--3926}.
\newblock


\bibitem[Loshchilov and Hutter(2016)]%
        {mg_28}
\bibfield{author}{\bibinfo{person}{Ilya Loshchilov} {and}
  \bibinfo{person}{Frank Hutter}.} \bibinfo{year}{2016}\natexlab{}.
\newblock \showarticletitle{Sgdr: Stochastic gradient descent with warm
  restarts}.
\newblock \bibinfo{journal}{\emph{arXiv preprint arXiv:1608.03983}}
  (\bibinfo{year}{2016}).
\newblock


\bibitem[Lu et~al\mbox{.}(2019)]%
        {indexnet}
\bibfield{author}{\bibinfo{person}{Hao Lu}, \bibinfo{person}{Yutong Dai},
  \bibinfo{person}{Chunhua Shen}, {and} \bibinfo{person}{Songcen Xu}.}
  \bibinfo{year}{2019}\natexlab{}.
\newblock \showarticletitle{Indices matter: Learning to index for deep image
  matting}. In \bibinfo{booktitle}{\emph{Proceedings of the IEEE/CVF
  International Conference on Computer Vision}}. \bibinfo{pages}{3266--3275}.
\newblock


\bibitem[Lutz et~al\mbox{.}(2018)]%
        {alphagan}
\bibfield{author}{\bibinfo{person}{Sebastian Lutz},
  \bibinfo{person}{Konstantinos Amplianitis}, {and} \bibinfo{person}{Aljosa
  Smolic}.} \bibinfo{year}{2018}\natexlab{}.
\newblock \showarticletitle{Alphagan: Generative adversarial networks for
  natural image matting}.
\newblock \bibinfo{journal}{\emph{arXiv preprint arXiv:1807.10088}}
  (\bibinfo{year}{2018}).
\newblock


\bibitem[Movahedi and Elder(2010)]%
        {sod_data}
\bibfield{author}{\bibinfo{person}{Vida Movahedi} {and}
  \bibinfo{person}{James~H Elder}.} \bibinfo{year}{2010}\natexlab{}.
\newblock \showarticletitle{Design and perceptual validation of performance
  measures for salient object segmentation}. In \bibinfo{booktitle}{\emph{2010
  IEEE computer society conference on computer vision and pattern
  recognition-workshops}}. IEEE, \bibinfo{pages}{49--56}.
\newblock


\bibitem[{NASA}(2015)]%
        {c:23}
\bibfield{author}{\bibinfo{person}{{NASA}}.} \bibinfo{year}{2015}\natexlab{}.
\newblock \bibinfo{title}{Pluto: The 'Other' Red Planet}.
\newblock
  \bibinfo{howpublished}{\url{https://www.nasa.gov/nh/pluto-the-other-red-planet}}.
\newblock
\newblock
\shownote{Accessed: 2018-12-06}.


\bibitem[Niu et~al\mbox{.}(2021)]%
        {harmony}
\bibfield{author}{\bibinfo{person}{Li Niu}, \bibinfo{person}{Wenyan Cong},
  \bibinfo{person}{Liu Liu}, \bibinfo{person}{Yan Hong}, \bibinfo{person}{Bo
  Zhang}, \bibinfo{person}{Jing Liang}, {and} \bibinfo{person}{Liqing Zhang}.}
  \bibinfo{year}{2021}\natexlab{}.
\newblock \showarticletitle{Making Images Real Again: A Comprehensive Survey on
  Deep Image Composition}.
\newblock \bibinfo{journal}{\emph{arXiv preprint arXiv:2106.14490}}
  (\bibinfo{year}{2021}).
\newblock


\bibitem[Pang et~al\mbox{.}(2020)]%
        {minet}
\bibfield{author}{\bibinfo{person}{Youwei Pang}, \bibinfo{person}{Xiaoqi Zhao},
  \bibinfo{person}{Lihe Zhang}, {and} \bibinfo{person}{Huchuan Lu}.}
  \bibinfo{year}{2020}\natexlab{}.
\newblock \showarticletitle{Multi-scale interactive network for salient object
  detection}. In \bibinfo{booktitle}{\emph{Proceedings of the IEEE/CVF
  Conference on Computer Vision and Pattern Recognition}}.
  \bibinfo{pages}{9413--9422}.
\newblock


\bibitem[Park et~al\mbox{.}(2022)]%
        {matteformer}
\bibfield{author}{\bibinfo{person}{GyuTae Park}, \bibinfo{person}{SungJoon
  Son}, \bibinfo{person}{JaeYoung Yoo}, \bibinfo{person}{SeHo Kim}, {and}
  \bibinfo{person}{Nojun Kwak}.} \bibinfo{year}{2022}\natexlab{}.
\newblock \showarticletitle{Matteformer: Transformer-based image matting via
  prior-tokens}. In \bibinfo{booktitle}{\emph{Proceedings of the IEEE/CVF
  Conference on Computer Vision and Pattern Recognition}}.
  \bibinfo{pages}{11696--11706}.
\newblock


\bibitem[Paszke et~al\mbox{.}(2017)]%
        {pytorch}
\bibfield{author}{\bibinfo{person}{Adam Paszke}, \bibinfo{person}{Sam Gross},
  \bibinfo{person}{Soumith Chintala}, \bibinfo{person}{Gregory Chanan},
  \bibinfo{person}{Edward Yang}, \bibinfo{person}{Zachary DeVito},
  \bibinfo{person}{Zeming Lin}, \bibinfo{person}{Alban Desmaison},
  \bibinfo{person}{Luca Antiga}, {and} \bibinfo{person}{Adam Lerer}.}
  \bibinfo{year}{2017}\natexlab{}.
\newblock \showarticletitle{Automatic differentiation in pytorch}.
\newblock  (\bibinfo{year}{2017}).
\newblock


\bibitem[Qiao et~al\mbox{.}(2020)]%
        {attention}
\bibfield{author}{\bibinfo{person}{Yu Qiao}, \bibinfo{person}{Yuhao Liu},
  \bibinfo{person}{Xin Yang}, \bibinfo{person}{Dongsheng Zhou},
  \bibinfo{person}{Mingliang Xu}, \bibinfo{person}{Qiang Zhang}, {and}
  \bibinfo{person}{Xiaopeng Wei}.} \bibinfo{year}{2020}\natexlab{}.
\newblock \showarticletitle{Attention-guided hierarchical structure aggregation
  for image matting}. In \bibinfo{booktitle}{\emph{Proceedings of the IEEE/CVF
  Conference on Computer Vision and Pattern Recognition}}.
  \bibinfo{pages}{13676--13685}.
\newblock


\bibitem[Qin(2020)]%
        {u2net}
\bibfield{author}{\bibinfo{person}{Xuebin Qin}.}
  \bibinfo{year}{2020}\natexlab{}.
\newblock \showarticletitle{U2-Net: Going deeper with nested U-structure for
  salient object detection}.
\newblock \bibinfo{journal}{\emph{Pattern Recognition}}  \bibinfo{volume}{106}
  (\bibinfo{year}{2020}), \bibinfo{pages}{107404}.
\newblock


\bibitem[Qin et~al\mbox{.}(2019)]%
        {basnet}
\bibfield{author}{\bibinfo{person}{Xuebin Qin}, \bibinfo{person}{Zichen Zhang},
  \bibinfo{person}{Chenyang Huang}, \bibinfo{person}{Chao Gao},
  \bibinfo{person}{Masood Dehghan}, {and} \bibinfo{person}{Martin Jagersand}.}
  \bibinfo{year}{2019}\natexlab{}.
\newblock \showarticletitle{Basnet: Boundary-aware salient object detection}.
  In \bibinfo{booktitle}{\emph{Proceedings of the IEEE/CVF Conference on
  Computer Vision and Pattern Recognition}}. \bibinfo{pages}{7479--7489}.
\newblock


\bibitem[Rice(1986)]%
        {r:86}
\bibfield{author}{\bibinfo{person}{James Rice}.}
  \bibinfo{year}{1986}\natexlab{}.
\newblock \bibinfo{booktitle}{\emph{{Poligon: A System for Parallel Problem
  Solving}}}.
\newblock \bibinfo{type}{Technical Report} KSL-86-19.
  \bibinfo{institution}{Dept.\ of Computer Science, Stanford Univ.}
\newblock


\bibitem[Robinson(1980a)]%
        {r:80}
\bibfield{author}{\bibinfo{person}{Arthur~L. Robinson}.}
  \bibinfo{year}{1980}\natexlab{a}.
\newblock \showarticletitle{New Ways to Make Microcircuits Smaller}.
\newblock \bibinfo{journal}{\emph{Science}} \bibinfo{volume}{208},
  \bibinfo{number}{4447} (\bibinfo{year}{1980}), \bibinfo{pages}{1019--1022}.
\newblock
\showISSN{0036-8075}
\urldef\tempurl%
\url{https://doi.org/10.1126/science.208.4447.1019}
\showDOI{\tempurl}
\showeprint{https://science.sciencemag.org/content/208/4447/1019.full.pdf}


\bibitem[Robinson(1980b)]%
        {r:80x}
\bibfield{author}{\bibinfo{person}{Arthur~L. Robinson}.}
  \bibinfo{year}{1980}\natexlab{b}.
\newblock \showarticletitle{{New Ways to Make Microcircuits Smaller---Duplicate
  Entry}}.
\newblock \bibinfo{journal}{\emph{Science}}  \bibinfo{volume}{208}
  (\bibinfo{year}{1980}), \bibinfo{pages}{1019--1026}.
\newblock


\bibitem[Ronneberger et~al\mbox{.}(2015)]%
        {unet}
\bibfield{author}{\bibinfo{person}{Olaf Ronneberger}, \bibinfo{person}{Philipp
  Fischer}, {and} \bibinfo{person}{Thomas Brox}.}
  \bibinfo{year}{2015}\natexlab{}.
\newblock \showarticletitle{U-net: Convolutional networks for biomedical image
  segmentation}. In \bibinfo{booktitle}{\emph{International Conference on
  Medical image computing and computer-assisted intervention}}. Springer,
  \bibinfo{pages}{234--241}.
\newblock


\bibitem[Sengupta et~al\mbox{.}(2020)]%
        {backv1}
\bibfield{author}{\bibinfo{person}{Soumyadip Sengupta}, \bibinfo{person}{Vivek
  Jayaram}, \bibinfo{person}{Brian Curless}, \bibinfo{person}{Steven~M Seitz},
  {and} \bibinfo{person}{Ira Kemelmacher-Shlizerman}.}
  \bibinfo{year}{2020}\natexlab{}.
\newblock \showarticletitle{Background matting: The world is your green
  screen}. In \bibinfo{booktitle}{\emph{Proceedings of the IEEE/CVF Conference
  on Computer Vision and Pattern Recognition}}. \bibinfo{pages}{2291--2300}.
\newblock


\bibitem[Shahrian et~al\mbox{.}(2013)]%
        {33}
\bibfield{author}{\bibinfo{person}{Ehsan Shahrian}, \bibinfo{person}{Deepu
  Rajan}, \bibinfo{person}{Brian Price}, {and} \bibinfo{person}{Scott Cohen}.}
  \bibinfo{year}{2013}\natexlab{}.
\newblock \showarticletitle{Improving image matting using comprehensive
  sampling sets}. In \bibinfo{booktitle}{\emph{Proceedings of the IEEE
  Conference on Computer Vision and Pattern Recognition}}.
  \bibinfo{pages}{636--643}.
\newblock


\bibitem[Shan et~al\mbox{.}(2021b)]%
        {shan2021uhrsnet}
\bibfield{author}{\bibinfo{person}{Lianlei Shan}, \bibinfo{person}{Minglong
  Li}, \bibinfo{person}{Xiaobin Li}, \bibinfo{person}{Yang Bai},
  \bibinfo{person}{Ke Lv}, \bibinfo{person}{Bin Luo}, \bibinfo{person}{Si-Bao
  Chen}, {and} \bibinfo{person}{Weiqiang Wang}.}
  \bibinfo{year}{2021}\natexlab{b}.
\newblock \showarticletitle{Uhrsnet: A semantic segmentation network
  specifically for ultra-high-resolution images}. In
  \bibinfo{booktitle}{\emph{2020 25th International Conference on Pattern
  Recognition (ICPR)}}. IEEE, \bibinfo{pages}{1460--1466}.
\newblock


\bibitem[Shan et~al\mbox{.}(2021a)]%
        {shan2021decouple}
\bibfield{author}{\bibinfo{person}{Lianlei Shan}, \bibinfo{person}{Xiaobin Li},
  {and} \bibinfo{person}{Weiqiang Wang}.} \bibinfo{year}{2021}\natexlab{a}.
\newblock \showarticletitle{Decouple the high-frequency and low-frequency
  information of images for semantic segmentation}. In
  \bibinfo{booktitle}{\emph{ICASSP 2021-2021 IEEE International Conference on
  Acoustics, Speech and Signal Processing (ICASSP)}}. IEEE,
  \bibinfo{pages}{1805--1809}.
\newblock


\bibitem[Shan and Wang(2021)]%
        {shan2021densenet}
\bibfield{author}{\bibinfo{person}{Lianlei Shan} {and}
  \bibinfo{person}{Weiqiang Wang}.} \bibinfo{year}{2021}\natexlab{}.
\newblock \showarticletitle{DenseNet-based land cover classification network
  with deep fusion}.
\newblock \bibinfo{journal}{\emph{IEEE Geoscience and Remote Sensing Letters}}
  \bibinfo{volume}{19} (\bibinfo{year}{2021}), \bibinfo{pages}{1--5}.
\newblock


\bibitem[Shan and Wang(2022)]%
        {shan2022mbnet}
\bibfield{author}{\bibinfo{person}{Lianlei Shan} {and}
  \bibinfo{person}{Weiqiang Wang}.} \bibinfo{year}{2022}\natexlab{}.
\newblock \showarticletitle{Mbnet: a multi-resolution branch network for
  semantic segmentation of ultra-high resolution images}. In
  \bibinfo{booktitle}{\emph{ICASSP 2022-2022 IEEE International Conference on
  Acoustics, Speech and Signal Processing (ICASSP)}}. IEEE,
  \bibinfo{pages}{2589--2593}.
\newblock


\bibitem[Shan et~al\mbox{.}(2021c)]%
        {shan2021class}
\bibfield{author}{\bibinfo{person}{Lianlei Shan}, \bibinfo{person}{Weiqiang
  Wang}, \bibinfo{person}{Ke Lv}, {and} \bibinfo{person}{Bin Luo}.}
  \bibinfo{year}{2021}\natexlab{c}.
\newblock \showarticletitle{Class-incremental learning for semantic
  segmentation in aerial imagery via distillation in all aspects}.
\newblock \bibinfo{journal}{\emph{IEEE Transactions on Geoscience and Remote
  Sensing}}  \bibinfo{volume}{60} (\bibinfo{year}{2021}),
  \bibinfo{pages}{1--12}.
\newblock


\bibitem[Shan et~al\mbox{.}(2022)]%
        {shan2022class}
\bibfield{author}{\bibinfo{person}{Lianlei Shan}, \bibinfo{person}{Weiqiang
  Wang}, \bibinfo{person}{Ke Lv}, {and} \bibinfo{person}{Bin Luo}.}
  \bibinfo{year}{2022}\natexlab{}.
\newblock \showarticletitle{Class-incremental semantic segmentation of aerial
  images via pixel-level feature generation and task-wise distillation}.
\newblock \bibinfo{journal}{\emph{IEEE Transactions on Geoscience and Remote
  Sensing}}  \bibinfo{volume}{60} (\bibinfo{year}{2022}),
  \bibinfo{pages}{1--17}.
\newblock


\bibitem[Shan et~al\mbox{.}(2023a)]%
        {shan2023boosting}
\bibfield{author}{\bibinfo{person}{Lianlei Shan}, \bibinfo{person}{Weiqiang
  Wang}, \bibinfo{person}{Ke Lv}, {and} \bibinfo{person}{Bin Luo}.}
  \bibinfo{year}{2023}\natexlab{a}.
\newblock \showarticletitle{Boosting Semantic Segmentation of Aerial Images via
  Decoupled and Multi-level Compaction and Dispersion}.
\newblock \bibinfo{journal}{\emph{IEEE Transactions on Geoscience and Remote
  Sensing}} (\bibinfo{year}{2023}).
\newblock


\bibitem[Shan et~al\mbox{.}(2023b)]%
        {shan2023data}
\bibfield{author}{\bibinfo{person}{Lianlei Shan}, \bibinfo{person}{Guiqin
  Zhao}, \bibinfo{person}{Jun Xie}, \bibinfo{person}{Peirui Cheng},
  \bibinfo{person}{Xiaobin Li}, {and} \bibinfo{person}{Zhepeng Wang}.}
  \bibinfo{year}{2023}\natexlab{b}.
\newblock \showarticletitle{A Data-Related Patch Proposal for Semantic
  Segmentation of Aerial Images}.
\newblock \bibinfo{journal}{\emph{IEEE Geoscience and Remote Sensing Letters}}
  \bibinfo{volume}{20} (\bibinfo{year}{2023}), \bibinfo{pages}{1--5}.
\newblock


\bibitem[Shan et~al\mbox{.}(2023c)]%
        {shan2023incremental}
\bibfield{author}{\bibinfo{person}{Leo Shan}, \bibinfo{person}{Wenzhang Zhou},
  {and} \bibinfo{person}{Grace Zhao}.} \bibinfo{year}{2023}\natexlab{c}.
\newblock \showarticletitle{Incremental Few Shot Semantic Segmentation via
  Class-agnostic Mask Proposal and Language-driven Classifier}. In
  \bibinfo{booktitle}{\emph{Proceedings of the 31st ACM International
  Conference on Multimedia}}. \bibinfo{pages}{8561--8570}.
\newblock


\bibitem[Sun et~al\mbox{.}(2004)]%
        {35}
\bibfield{author}{\bibinfo{person}{Jian Sun}, \bibinfo{person}{Jiaya Jia},
  \bibinfo{person}{Chi-Keung Tang}, {and} \bibinfo{person}{Heung-Yeung Shum}.}
  \bibinfo{year}{2004}\natexlab{}.
\newblock \showarticletitle{Poisson matting}.
\newblock In \bibinfo{booktitle}{\emph{ACM SIGGRAPH 2004 Papers}}.
  \bibinfo{pages}{315--321}.
\newblock


\bibitem[Sun et~al\mbox{.}(2021a)]%
        {semantic}
\bibfield{author}{\bibinfo{person}{Yanan Sun} {et~al\mbox{.}}}
  \bibinfo{year}{2021}\natexlab{a}.
\newblock \showarticletitle{Semantic Image Matting}. In
  \bibinfo{booktitle}{\emph{Proceedings of the IEEE/CVF Conference on Computer
  Vision and Pattern Recognition}}. \bibinfo{pages}{11120--11129}.
\newblock


\bibitem[Sun et~al\mbox{.}(2021b)]%
        {video_matting_dataset}
\bibfield{author}{\bibinfo{person}{Yanan Sun}, \bibinfo{person}{Guanzhi Wang},
  \bibinfo{person}{Qiao Gu}, \bibinfo{person}{Chi-Keung Tang}, {and}
  \bibinfo{person}{Yu-Wing Tai}.} \bibinfo{year}{2021}\natexlab{b}.
\newblock \showarticletitle{Deep Video Matting via Spatio-Temporal Alignment
  and Aggregation}. In \bibinfo{booktitle}{\emph{Proceedings of the IEEE/CVF
  Conference on Computer Vision and Pattern Recognition}}.
  \bibinfo{pages}{6975--6984}.
\newblock


\bibitem[Tang(2019)]%
        {learning_based}
\bibfield{author}{\bibinfo{person}{Aksoy Yagiz-Oztireli Cengiz Gross Markus
  Aydin Tunc~Ozan Tang, Jingwei}.} \bibinfo{year}{2019}\natexlab{}.
\newblock \showarticletitle{Learning-based sampling for natural image matting}.
  In \bibinfo{booktitle}{\emph{Proceedings of the IEEE/CVF Conference on
  Computer Vision and Pattern Recognition}}. \bibinfo{pages}{3055--3063}.
\newblock


\bibitem[Tian et~al\mbox{.}(2020a)]%
        {condinst}
\bibfield{author}{\bibinfo{person}{Zhi Tian}, \bibinfo{person}{Chunhua Shen},
  {and} \bibinfo{person}{Hao Chen}.} \bibinfo{year}{2020}\natexlab{a}.
\newblock \showarticletitle{Conditional convolutions for instance
  segmentation}. In \bibinfo{booktitle}{\emph{Computer Vision--ECCV 2020: 16th
  European Conference, Glasgow, UK, August 23--28, 2020, Proceedings, Part I
  16}}. Springer, \bibinfo{pages}{282--298}.
\newblock


\bibitem[Tian et~al\mbox{.}(2020b)]%
        {solov1}
\bibfield{author}{\bibinfo{person}{Zhi Tian}, \bibinfo{person}{Chunhua Shen},
  {and} \bibinfo{person}{Hao Chen}.} \bibinfo{year}{2020}\natexlab{b}.
\newblock \showarticletitle{Conditional convolutions for instance
  segmentation}. In \bibinfo{booktitle}{\emph{Computer Vision--ECCV 2020: 16th
  European Conference, Glasgow, UK, August 23--28, 2020, Proceedings, Part I
  16}}. Springer, \bibinfo{pages}{282--298}.
\newblock


\bibitem[Ulyanov et~al\mbox{.}(2016)]%
        {ins_norm}
\bibfield{author}{\bibinfo{person}{Dmitry Ulyanov}, \bibinfo{person}{Andrea
  Vedaldi}, {and} \bibinfo{person}{Victor Lempitsky}.}
  \bibinfo{year}{2016}\natexlab{}.
\newblock \showarticletitle{Instance normalization: The missing ingredient for
  fast stylization}.
\newblock \bibinfo{journal}{\emph{arXiv preprint arXiv:1607.08022}}
  (\bibinfo{year}{2016}).
\newblock


\bibitem[Wang et~al\mbox{.}(2020)]%
        {pfpn}
\bibfield{author}{\bibinfo{person}{Bo Wang}, \bibinfo{person}{Quan Chen},
  \bibinfo{person}{Min Zhou}, \bibinfo{person}{Zhiqiang Zhang},
  \bibinfo{person}{Xiaogang Jin}, {and} \bibinfo{person}{Kun Gai}.}
  \bibinfo{year}{2020}\natexlab{}.
\newblock \showarticletitle{Progressive feature polishing network for salient
  object detection}. In \bibinfo{booktitle}{\emph{Proceedings of the AAAI
  Conference on Artificial Intelligence}}, Vol.~\bibinfo{volume}{34}.
  \bibinfo{pages}{12128--12135}.
\newblock


\bibitem[Wang and Cohen(2007a)]%
        {survey}
\bibfield{author}{\bibinfo{person}{Jue Wang} {and} \bibinfo{person}{Michael~F
  Cohen}.} \bibinfo{year}{2007}\natexlab{a}.
\newblock \bibinfo{booktitle}{\emph{Image and video matting}}.
\newblock \bibinfo{publisher}{Now Publishers}.
\newblock


\bibitem[Wang and Cohen(2007b)]%
        {38}
\bibfield{author}{\bibinfo{person}{Jue Wang} {and} \bibinfo{person}{Michael~F
  Cohen}.} \bibinfo{year}{2007}\natexlab{b}.
\newblock \showarticletitle{Optimized color sampling for robust matting}. In
  \bibinfo{booktitle}{\emph{2007 IEEE Conference on Computer Vision and Pattern
  Recognition}}. IEEE, \bibinfo{pages}{1--8}.
\newblock


\bibitem[Wang et~al\mbox{.}(2017)]%
        {duts}
\bibfield{author}{\bibinfo{person}{Lijun Wang}, \bibinfo{person}{Huchuan Lu},
  \bibinfo{person}{Yifan Wang}, \bibinfo{person}{Mengyang Feng},
  \bibinfo{person}{Dong Wang}, \bibinfo{person}{Baocai Yin}, {and}
  \bibinfo{person}{Xiang Ruan}.} \bibinfo{year}{2017}\natexlab{}.
\newblock \showarticletitle{Learning to Detect Salient Objects with Image-level
  Supervision}. In \bibinfo{booktitle}{\emph{CVPR}}.
\newblock


\bibitem[Wei et~al\mbox{.}(2020)]%
        {f3net}
\bibfield{author}{\bibinfo{person}{Jun Wei}, \bibinfo{person}{Shuhui Wang},
  {and} \bibinfo{person}{Qingming Huang}.} \bibinfo{year}{2020}\natexlab{}.
\newblock \showarticletitle{F$^3$Net: Fusion, Feedback and Focus for Salient
  Object Detection}. In \bibinfo{booktitle}{\emph{Proceedings of the AAAI
  Conference on Artificial Intelligence}}, Vol.~\bibinfo{volume}{34}.
  \bibinfo{pages}{12321--12328}.
\newblock


\bibitem[Wu et~al\mbox{.}(2023)]%
        {wu2023continual}
\bibfield{author}{\bibinfo{person}{Weijia Wu}, \bibinfo{person}{Yuzhong Zhao},
  \bibinfo{person}{Zhuang Li}, \bibinfo{person}{Lianlei Shan},
  \bibinfo{person}{Hong Zhou}, {and} \bibinfo{person}{Mike~Zheng Shou}.}
  \bibinfo{year}{2023}\natexlab{}.
\newblock \showarticletitle{Continual Learning for Image Segmentation with
  Dynamic Query}.
\newblock \bibinfo{journal}{\emph{IEEE Transactions on Circuits and Systems for
  Video Technology}} (\bibinfo{year}{2023}).
\newblock


\bibitem[Xie and Tu(2015)]%
        {hed}
\bibfield{author}{\bibinfo{person}{Saining Xie} {and} \bibinfo{person}{Zhuowen
  Tu}.} \bibinfo{year}{2015}\natexlab{}.
\newblock \showarticletitle{Holistically-nested edge detection}. In
  \bibinfo{booktitle}{\emph{Proceedings of the IEEE international conference on
  computer vision}}. \bibinfo{pages}{1395--1403}.
\newblock


\bibitem[Xu et~al\mbox{.}(2017)]%
        {dim}
\bibfield{author}{\bibinfo{person}{Ning Xu}, \bibinfo{person}{Brian Price},
  \bibinfo{person}{Scott Cohen}, {and} \bibinfo{person}{Thomas Huang}.}
  \bibinfo{year}{2017}\natexlab{}.
\newblock \showarticletitle{Deep image matting}. In
  \bibinfo{booktitle}{\emph{Proceedings of the IEEE conference on computer
  vision and pattern recognition}}. \bibinfo{pages}{2970--2979}.
\newblock


\bibitem[Yu et~al\mbox{.}(2020)]%
        {mgmatting}
\bibfield{author}{\bibinfo{person}{Qihang Yu}, \bibinfo{person}{Jianming
  Zhang}, \bibinfo{person}{He Zhang}, \bibinfo{person}{Yilin Wang},
  \bibinfo{person}{Zhe Lin}, \bibinfo{person}{Ning Xu}, \bibinfo{person}{Yutong
  Bai}, {and} \bibinfo{person}{Alan Yuille}.} \bibinfo{year}{2020}\natexlab{}.
\newblock \showarticletitle{Mask Guided Matting via Progressive Refinement
  Network}.
\newblock \bibinfo{journal}{\emph{arXiv e-prints}} (\bibinfo{year}{2020}),
  \bibinfo{pages}{arXiv--2012}.
\newblock


\bibitem[Zhang et~al\mbox{.}(2019)]%
        {lfm}
\bibfield{author}{\bibinfo{person}{Yunke Zhang}, \bibinfo{person}{Lixue Gong},
  \bibinfo{person}{Lubin Fan}, \bibinfo{person}{Peiran Ren},
  \bibinfo{person}{Qixing Huang}, \bibinfo{person}{Hujun Bao}, {and}
  \bibinfo{person}{Weiwei Xu}.} \bibinfo{year}{2019}\natexlab{}.
\newblock \showarticletitle{A late fusion cnn for digital matting}. In
  \bibinfo{booktitle}{\emph{Proceedings of the IEEE/CVF Conference on Computer
  Vision and Pattern Recognition}}. \bibinfo{pages}{7469--7478}.
\newblock


\bibitem[Zhao et~al\mbox{.}(2017)]%
        {pspnet}
\bibfield{author}{\bibinfo{person}{Hengshuang Zhao}, \bibinfo{person}{Jianping
  Shi}, \bibinfo{person}{Xiaojuan Qi}, \bibinfo{person}{Xiaogang Wang}, {and}
  \bibinfo{person}{Jiaya Jia}.} \bibinfo{year}{2017}\natexlab{}.
\newblock \showarticletitle{Pyramid scene parsing network}. In
  \bibinfo{booktitle}{\emph{Proceedings of the IEEE conference on computer
  vision and pattern recognition}}. \bibinfo{pages}{2881--2890}.
\newblock


\bibitem[Zhao et~al\mbox{.}(2019)]%
        {egnet}
\bibfield{author}{\bibinfo{person}{Jia-Xing Zhao}, \bibinfo{person}{Jiang-Jiang
  Liu}, \bibinfo{person}{Deng-Ping Fan}, \bibinfo{person}{Yang Cao},
  \bibinfo{person}{Jufeng Yang}, {and} \bibinfo{person}{Ming-Ming Cheng}.}
  \bibinfo{year}{2019}\natexlab{}.
\newblock \showarticletitle{EGNet: Edge guidance network for salient object
  detection}. In \bibinfo{booktitle}{\emph{Proceedings of the IEEE/CVF
  International Conference on Computer Vision}}. \bibinfo{pages}{8779--8788}.
\newblock


\end{thebibliography}
\appendix
\end{document}